\documentclass[11pt]{article}

\usepackage[preprint]{acl}

\usepackage{times}
\usepackage{latexsym}

\usepackage[T1]{fontenc}

\usepackage[utf8]{inputenc}

\usepackage{microtype}

\usepackage{inconsolata}

\usepackage{graphicx}

\usepackage{amsmath}
\usepackage{amssymb}
\usepackage{booktabs}
\usepackage{tabularx}
\usepackage{pifont}
\usepackage{needspace}
\usepackage{placeins}
\definecolor{comparisonTeal}{HTML}{008C8C}
\definecolor{comparisonRose}{HTML}{C5004F}
\definecolor{comparisonNavy}{HTML}{000099}
\hypersetup{colorlinks=true,citecolor=comparisonNavy,linkcolor=comparisonNavy,urlcolor=comparisonNavy}
\newcommand{\tableyes}{\textcolor{comparisonTeal}{\ding{51}}}
\newcommand{\tableno}{\textcolor{comparisonRose}{\ding{55}}}

\title{Small Edits, Shifting Decisions: Benchmarking LLM Accuracy\newline and Reliability on French Input Variants}
\author{
  Jiaming Yu$^{1}$ \quad Xiaotian Xie$^{2}$ \quad Jian Ma$^{1,3}$ \quad Yingjie Li$^{1}$\thanks{Corresponding author.} \\
  $^{1}$Hangzhou International Innovation Institute, Beihang University, Hangzhou, China \\
  $^{2}$School of Automation, Central South University, Changsha, China \\
  $^{3}$School of Reliability and Systems Engineering, Beihang University, China \\
  \texttt{\{yjmsoleilune,yingjieli\}@buaa.edu.cn}
}

\begin{document}
\maketitle

\begin{abstract}
French users may omit apostrophes and diacritics or make keyboard typos in everyday input. The effects of these common variations on language model performance and prediction reliability remain insufficiently understood. To address this gap, we construct a controlled French surface-variant benchmark covering seven task types, with 4,200 original--low--high triplets (12,600 input instances). Each high-intensity variant retains all low-intensity edits and adds further changes. We evaluate six models from two LLM families. The results show that accuracy decreases and Brier score and negative log-likelihood (NLL) increase with perturbation intensity. Larger models show smaller accuracy losses under high intensity, but increases in Brier score and NLL do not consistently decrease with scale. Keyboard typos cause the largest losses in accuracy and the quality of predicted probabilities at the predefined high-intensity settings. Finally, calibration temperatures fitted on original inputs transfer effectively to surface variants. Combining calibration with input restoration further improves Brier score and NLL over either method alone while preserving restored accuracy. Our benchmark provides a unified framework for jointly evaluating task accuracy and prediction reliability under common French surface variations.
\end{abstract}

\section{Introduction}
\label{sec:introduction}
\begin{figure}[!t]
\centering
\includegraphics[width=\columnwidth,trim=0 0 0 6bp,clip]{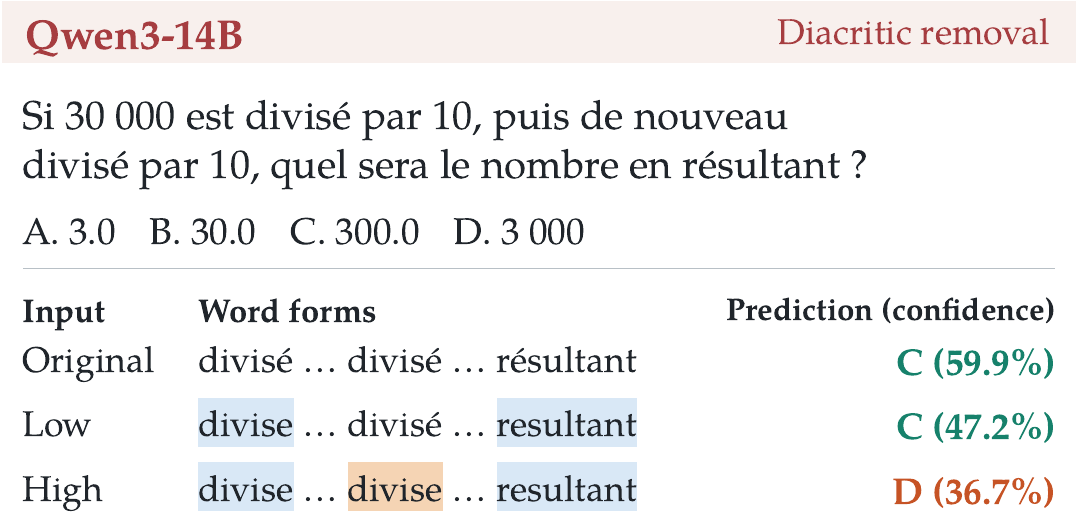}
\caption{Diacritic removal on a question asking for $30{,}000/10/10$. Qwen3-14B retains the correct answer C (300) at low perturbation intensity and predicts D at high perturbation intensity. Percentages give the probability of the selected answer. Blue marks shared edits; orange marks the additional high-intensity edit. Numbers, options, and omitted text are unchanged.}
\label{fig:case}
\end{figure}

Large language models (LLMs) are widely used in applications such as question answering and conversational systems \citep{naveed2025overview}. However, French inputs in everyday use do not always follow standard written conventions. Users may omit apostrophes or diacritics or make keyboard typos, e.g., writing ``jai'' instead of ``j'ai'' or ``ecole'' instead of ``école'' \citep{fairon-paumier-2006-translated,vancompernolle-2011-diacritics,pinet-etal-2022-typing}. Although these forms may remain understandable from context, small changes can affect model predictions. Figure~\ref{fig:case} illustrates a prediction flip following diacritic removal, with the probability of the correct answer also changing before the prediction flips. Such changes may not be fully captured by accuracy measured on original French inputs alone. Understanding how these variations affect both task performance and prediction reliability is therefore important for evaluating LLMs in French-language applications.

\begin{table*}[t]
\centering
\small
\setlength{\tabcolsep}{3pt}
\begin{tabular*}{\textwidth}{@{\extracolsep{\fill}}lllccccc@{}}
\toprule
Name & Authors & Input & Fr. & \multicolumn{2}{c}{Perturbations} & \multicolumn{2}{c}{Evaluation} \\
\cmidrule(lr){5-6}\cmidrule(lr){7-8}
 & & & & Paired & Levels & Task & Prob. \\
\midrule
FLUE & \textcolor{comparisonNavy}{\citet{le-etal-2020-flaubert-unsupervised}} & Original & \tableyes & \tableno & \tableno & \tableyes & \tableno \\
FQuAD & \textcolor{comparisonNavy}{\citet{dhoffschmidt-etal-2020-fquad}} & Original & \tableyes & \tableno & \tableno & \tableyes & \tableno \\
FrenchBench & \textcolor{comparisonNavy}{\citet{faysse2025}} & Original & \tableyes & \tableno & \tableno & \tableyes & \tableno \\
\midrule
MTNT & \textcolor{comparisonNavy}{\citet{michel-neubig-2018-mtnt}} & Natural noise & \tableyes & \tableno & \tableno & \tableyes & \tableno \\
Noised tests & \textcolor{comparisonNavy}{\citet{cooper-stickland-etal-2023-robustification}} & Injected errors & \tableyes & \tableyes & \tableno & \tableyes & \tableno \\
Noisy text & \textcolor{comparisonNavy}{\citet{zhang-etal-2023-transferable}} & Text edits & \tableno & \tableyes & \tableyes & \tableyes & \tableyes \\
Ad-Word & \textcolor{comparisonNavy}{\citet{moffett-dhingra-2025-close}} & Adversarial & \tableno & \tableyes & \tableyes & \tableyes & \tableno \\
MULTYPO & \textcolor{comparisonNavy}{\citet{zhao-etal-2026-evaluating}} & Keyboard typos & \tableyes & \tableyes & \tableyes & \tableyes & \tableno \\
\midrule
\textbf{This work} & \textbf{Ours} & Three variations & \tableyes & \tableyes & \tableyes & \tableyes & \tableyes \\
\bottomrule
\end{tabular*}
\caption{Comparison of French benchmarks and text-robustness evaluations. Our work considers three French surface variations: apostrophe omission, diacritic removal, and keyboard typos. Fr. indicates French coverage. Paired denotes matched original and perturbed inputs. Levels denotes evaluation at multiple perturbation intensities. Task denotes evaluation of downstream task performance using task-specific metrics. Prob. denotes evaluation of probability-based reliability using calibration or proper scoring metrics.}
\label{tab:benchmark-comparison}
\vspace{-2mm}
\end{table*}

French NLP evaluation has advanced through benchmarks such as FLUE, introduced alongside FlauBERT \citep{le-etal-2020-flaubert-unsupervised}, and FrenchBench, which extends evaluation to generative language models \citep{faysse2025}. Previous robustness studies have examined French machine translation and multilingual task performance under natural or injected text noise \citep{michel-neubig-2018-mtnt,cooper-stickland-etal-2023-robustification}. More recent work has evaluated LLMs on noisy instructions \citep{wang-etal-2024-resilience} and language-specific keyboard typos across multiple languages, including French \citep{zhao-etal-2026-evaluating}. 
Related work has also investigated temperature calibration under noisy text classification \citep{zhang-etal-2023-transferable} and the role of word recovery in robustness to adversarial text perturbations \citep{moffett-dhingra-2025-close}.

Despite these advances, two gaps remain in the evaluation of French surface variations. First, existing French benchmarks primarily evaluate original task inputs \citep{le-etal-2020-flaubert-unsupervised,faysse2025}, while robustness studies consider different types of noisy text \citep{wang-etal-2024-resilience,peters-martins-2025-translation,zhao-etal-2026-evaluating}. Systematic comparisons of apostrophe omission, diacritic removal, and keyboard typos across the same French tasks, using paired inputs and nested perturbation intensities, remain limited. Second, changes in task accuracy do not fully capture changes in predicted probabilities. Although calibration under noisy text has been investigated \citep{zhang-etal-2023-transferable}, how these French surface variations jointly affect task performance and prediction reliability remains less well understood.

To address these gaps, we construct a French surface-variant benchmark covering seven task types and the three writing variations: apostrophe omission, diacritic removal, and keyboard typos. Each original input is paired with low- and high-intensity variants, with the high-intensity variant retaining all low-intensity edits and adding further changes. This design allows us to examine the effects of increasing perturbation on the same input. We evaluate six models from the Qwen3 and Gemma-3 families, measuring both answer correctness and the quality of predicted probabilities. We further investigate prediction flips in relation to original decision margins and examine confidence--accuracy relationships across input conditions. Finally, we assess the transferability of calibration fitted on original inputs and the complementary effects of input restoration. Table~\ref{tab:benchmark-comparison} compares our evaluation with prior work.

Our main contributions are as follows. (i) We construct a controlled French surface-variant benchmark containing 4,200 triplets (12,600 inputs), covering seven task types, three writing variations, and two nested perturbation intensities. (ii) We systematically evaluate task performance and prediction reliability across models, perturbation types, and intensities. We find that increasing perturbation intensity consistently degrades accuracy and the quality of predicted probabilities. Moreover, larger models exhibit greater stability in accuracy without consistently smaller degradation in the quality of predicted probabilities. Among the three variations, keyboard typos have the strongest effects at the predefined high-intensity settings. (iii) We empirically show that calibration temperatures fitted on original inputs transfer effectively to surface variants. Combining calibration with input restoration yields the lowest Brier score and NLL among the evaluated methods while preserving restored accuracy.

\section{Related Work}
\label{sec:related-work}
\textbf{French NLP and Evaluation.} Research on French natural language processing (NLP) has established dedicated language models and multitask evaluation resources. CamemBERT and FlauBERT advanced French language representations \citep{martin-etal-2020-camembert,le-etal-2020-flaubert-unsupervised}, with FlauBERT also introducing FLUE. FQuAD and OrangeSum provide data for reading comprehension and summarization, while BARThez provides a pretrained French sequence-to-sequence model \citep{dhoffschmidt-etal-2020-fquad,eddine-etal-2021-barthez}. More recently, \citet{faysse2025} introduced CroissantLLM and FrenchBench for French LLM evaluation. FLUE, FQuAD, and FrenchBench broaden task coverage but do not provide matched evaluations of controlled French surface variations.

\textbf{Surface Variation in French Writing.} Corpus and behavioral studies have documented surface-form variation in digitally mediated French writing. \citet{fairon-paumier-2006-translated} discussed apostrophe omission and restoration in the normalization of French text messages. \citet{vancompernolle-2011-diacritics} examined French diacritic use on an online dating site and found sociolinguistic variation in their retention and omission. \citet{pinet-etal-2022-typing} investigated typing errors through typing tasks and a survey of native French speakers' typing habits. These studies motivate our choice of apostrophe omission, diacritic removal, and keyboard typos.

\textbf{Robustness and Reliability.} The French Social Media Bank and Machine Translation of Noisy Text (MTNT) examined French social media processing and translation under naturally noisy inputs \citep{seddah-etal-2012-french,michel-neubig-2018-mtnt}. \citet{cooper-stickland-etal-2023-robustification} evaluated tasks in French and other languages using injected errors derived from real-world writing. For LLMs, \citet{gan-etal-2024-reasoning} examined reasoning under adversarial typographical errors, while MULTYPO compared language-specific keyboard typos at different perturbation intensities \citep{zhao-etal-2026-evaluating}. These studies mainly evaluate task performance under natural noise or typographical errors. How different French surface variations affect both accuracy and prediction reliability remains less clear.

Previous work has studied confidence calibration in pretrained language models, question answering, and multilingual settings \citep{desai-durrett-2020-calibration,jiang-etal-2021-know,jiang-etal-2022-calibrating}. Under distribution shifts and noisy inputs, calibration adjustments have been explored by \citet{zhang-etal-2023-transferable} and \citet{gasparin-detommaso-2024-distance}. Complementary efforts have investigated input correction and its effects on downstream tasks \citep{pruthi-etal-2019-combating,hanneman-etal-2024-impacts,moffett-dhingra-2025-close}. Together, these studies advance French language evaluation and research on noisy inputs, but the joint effects of French surface variations on accuracy and prediction reliability remain unclear. Our paired benchmark examines these effects and the complementary benefits of calibration and restoration.

\section{Dataset Construction}
\label{sec:dataset}

\subsection{Data Sources and Selection}
\label{sec:data-sources}
We select datasets from existing French resources that cover seven task types, including the Multilingual Amazon Reviews Corpus (MARC), FrenchNLI, Belebele, French HellaSwag, the French mathematics subset of Global-MMLU, HistoiresMorales, and Multilingual and Cross-lingual Word-in-Context Disambiguation (MCL-WiC). These tasks have well-defined candidate answers, allowing us to examine both answer correctness and predicted probabilities under a unified framework.
Sample selection considers task requirements and the availability of enough editable words to support two levels of nested perturbations. For each dataset, we select 200 original questions for each of the three perturbation factors, with non-overlapping original question sets for different factors. The resulting benchmark contains 4,200 original--low--high triplets, corresponding to 12,600 input instances. The tasks, number of candidate answers, and selection sizes for each dataset are shown in Table~\ref{tab:data-sources}.

\begin{table*}[t]
\centering
\small
\setlength{\tabcolsep}{5pt}
\renewcommand{\arraystretch}{1.00}
\begin{tabularx}{\textwidth}{@{}p{0.40\textwidth}Xcc@{}}
\toprule
Dataset & Task & \#Choices & \#Triplets \\
\midrule
MARC \citep{keung-etal-2020-multilingual} & Review rating & 5 & 600 \\
\href{https://huggingface.co/datasets/CATIE-AQ/frenchNLI}{FrenchNLI} & Natural language inference (NLI) & 3 & 600 \\
Belebele \citep{bandarkar-etal-2024-belebele} & Reading comprehension & 4 & 600 \\
HellaSwag \citep{zellers-etal-2019-hellaswag,faysse2025} & Commonsense continuation & 4 & 600 \\
Global-MMLU \citep{singh-etal-2025-global} & Mathematical reasoning & 4 & 600 \\
HistoiresMorales \citep{leteno-etal-2025-histoiresmorales} & Moral action judgment & 2 & 600 \\
MCL-WiC \citep{martelli-etal-2021-semeval} & Word-in-context judgment & 2 & 600 \\
\midrule
Total & & & 4,200 \\
\bottomrule
\end{tabularx}
\caption{Source datasets and selected triplets. All source inputs are in French; Global-MMLU is restricted to mathematics. Each base item forms one original--low--high triplet.}
\label{tab:data-sources}
\end{table*}

\subsection{Perturbation Construction}
\label{sec:perturbation-construction}
\begin{figure*}[t]
\centering
\includegraphics[width=\textwidth,trim=0 0 0 8bp,clip]{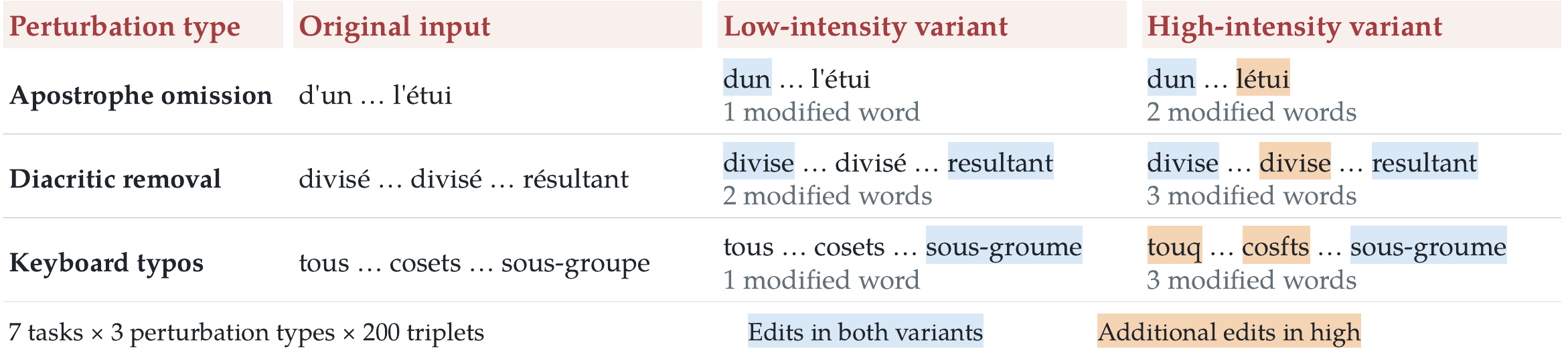}
\caption{Construction of low- and high-intensity variants for the three perturbation types. Rows show selected original words and their variants; English glosses are in Appendix~\ref{app:example-glosses}. Blue marks shared edits; orange marks additional high-intensity edits. Counts give modified words; ellipses indicate omitted context.}
\label{fig:construction}
\end{figure*}

We convert the datasets to a unified format for candidate answer evaluation and separately apply apostrophe omission, diacritic removal, and keyboard typos to the original inputs. Apostrophe omission targets French elision forms that meet predefined rules; diacritic removal eliminates diacritical marks from selected words; keyboard typos are generated by replacing characters with neighboring keys on an AZERTY keyboard.

As illustrated in Figure~\ref{fig:construction}, for each original question, we generate two levels of variants: low and high. The target modification proportions for apostrophe omission are 2.5\% and 5\%, while for diacritic removal and keyboard typos, they are 5\% and 10\%. We use lower target proportions for apostrophe omission because fewer words contain eligible French elision forms. The modification proportion is the fraction of original input words modified. We compute edit counts from the word count and target proportion using random rounding with deterministic seeds, so actual proportions may differ slightly. The two levels of variants share the same perturbation plan: high retains all modifications from low and adds at least one additional modified word. Candidate samples that cannot meet the two levels of intensity and strict nesting requirements are excluded from selection.

To preserve task integrity, we fix the reference answers and option order across all versions of the same original question, and protect numerical values, structural task markers, and target words in contextual word sense tasks. Predefined lexical collisions are excluded during construction, while remaining word-form overlaps are assessed in context during quality review. The perturbation rules and protection scopes are provided in Appendix~\ref{app:perturbation-rules}; mean realized perturbation proportions are reported in Table~\ref{tab:benchmark-statistics}.

\subsection{Quality Validation}
\label{sec:quality-validation}
DeepSeek-V4-Pro and Gemini-2.5-Pro independently screened the triplets for preservation of task-relevant information and reference-answer validity. Each model received the original input, both variants, edit records, answer options, and the reference answer. It first checked the original question and its reference answer, then assessed each variant against the original, returning separate judgments and brief explanations. The screening instructions and decision rules are provided in Appendix~\ref{app:automatic-review}.
We included 26 triplets with failed or invalid Gemini assessments in targeted human review and randomly sampled 210 triplets from the remaining samples for a stratified audit, with 10 per dataset and perturbation factor combination, covering 5\% of all triplets.

The human review was conducted by Lan-bridge,\footnote{\url{https://www.lan-bridge.com/enus}} a professional language service provider specializing in translation and localization. All three annotators held French proficiency at level B2 or above on the Common European Framework of Reference for Languages (CEFR), and were provided with standardized guidelines alongside annotated examples. The three annotators independently evaluated the low- and high-intensity variants in each triplet, resolving discrepancies via majority vote. On the random audit batch, Krippendorff's $\alpha$ \citep{artstein-poesio-2008-survey} was 0.969 for the initial ratings and reached 1.000 for the final accepted ratings after clarification of the annotation guidelines. Following human review and source-question verification, we replaced four triplets, with all replacements subsequently approved by both screening models. Complete review rubrics and outcomes are detailed in Appendix~\ref{app:human-review}.

\subsection{Dataset Splits and Statistics}
\label{sec:dataset-splits}
We partitioned the 4,200 triplets into a held-out set of 840 triplets and a test set of 3,360 triplets, providing a dedicated resource for validation and final evaluation, respectively. In this work, post-hoc calibration temperatures were fitted exclusively on the held-out set. Each original question and its two variants were always placed in the same split. Furthermore, for samples that share passages or source groups, partitioning was conducted at the source-group level, precluding overlap across perturbation types or evaluation splits. Table~\ref{tab:benchmark-statistics} summarizes the split statistics and realized perturbation intensities.

\begin{table}[t]
\centering\small
\setlength{\tabcolsep}{3pt}
\begin{tabularx}{\columnwidth}{@{}Xcccc@{}}
\toprule
 & \multicolumn{2}{c}{\#Triplets} & \multicolumn{2}{c}{Mean edits (\%)} \\
\cmidrule(lr){2-3}\cmidrule(lr){4-5}
Perturbation & Held-out & Test & Low & High \\
\midrule
Apostrophe omission & 280 & 1,120 & 2.50 & 5.10 \\
Diacritic removal & 280 & 1,120 & 4.99 & 9.96 \\
Keyboard typos & 280 & 1,120 & 5.00 & 10.14 \\
\midrule
Total & 840 & 3,360 & -- & -- \\
\bottomrule
\end{tabularx}
\caption{Statistics of the final benchmark: 200 base items $\times$ 7 datasets $\times$ 3 perturbation types $=$ 4,200 triplets. Each triplet contains Original, Low, and High inputs, giving $4{,}200\times3=12{,}600$ input instances. Each dataset contributes 40 held-out and 160 test triplets per perturbation type. Mean edits are the average proportions of original words modified, computed across both splits for each intensity.}
\label{tab:benchmark-statistics}
\end{table}

\begin{table*}[t]
\centering
\small
\setlength{\tabcolsep}{2.5pt}
\renewcommand{\arraystretch}{1.00}
\begin{tabular*}{\textwidth}{@{\extracolsep{\fill}}l*{12}{c}@{}}
\toprule
 & \multicolumn{4}{c}{Original} & \multicolumn{4}{c}{Low} & \multicolumn{4}{c}{High} \\
\cmidrule(lr){2-5}\cmidrule(lr){6-9}\cmidrule(lr){10-13}
Model & Acc.$\uparrow$ & ECE$\downarrow$ & Brier$\downarrow$ & NLL$\downarrow$ & Acc.$\uparrow$ & ECE$\downarrow$ & Brier$\downarrow$ & NLL$\downarrow$ & Acc.$\uparrow$ & ECE$\downarrow$ & Brier$\downarrow$ & NLL$\downarrow$ \\
\midrule
Qwen3-8B & 73.96 & 4.67 & 0.370 & 0.654 & 73.18 & 4.88 & 0.379\rlap{\textsuperscript{*}} & 0.668\rlap{\textsuperscript{*}} & 72.17\rlap{\textsuperscript{*}} & 5.61 & 0.388\rlap{\textsuperscript{*}} & 0.681\rlap{\textsuperscript{*}} \\
Qwen3-14B & 78.39 & 6.49 & 0.314 & 0.571 & 78.01 & 5.82 & 0.318 & 0.579 & 77.08\rlap{\textsuperscript{*}} & 6.02 & 0.323\rlap{\textsuperscript{*}} & 0.589\rlap{\textsuperscript{*}} \\
Qwen3-32B & 80.65 & 3.77 & 0.276 & 0.506 & 80.48 & 4.20 & 0.283\rlap{\textsuperscript{*}} & 0.520\rlap{\textsuperscript{*}} & 79.97 & 4.57 & 0.289\rlap{\textsuperscript{*}} & 0.529\rlap{\textsuperscript{*}} \\
\midrule
Gemma-3-4B-it & 62.41 & 23.95 & 0.597 & 1.123 & 61.67 & 24.09 & 0.603 & 1.127 & 61.07 & 24.19 & 0.608 & 1.130 \\
Gemma-3-12B-it & 75.86 & 16.41 & 0.394 & 0.789 & 75.68 & 16.17 & 0.399 & 0.796 & 74.70 & 16.76 & 0.407 & 0.806 \\
Gemma-3-27B-it & 78.66 & 16.44 & 0.368 & 0.873 & 78.07 & 16.69 & 0.377 & 0.891 & 78.04 & 16.63 & 0.379 & 0.897 \\
\bottomrule
\end{tabular*}
\caption{Overall performance on the 3,360 test triplets from our internal split, reported separately for each condition across seven datasets and three perturbation factors. Acc. denotes accuracy and Brier denotes Brier score. Acc. and ECE are percentages; Brier and NLL retain their original scales. Probabilities are uncalibrated. $^{*}$ marks a difference from Original at Holm-adjusted $p<0.05$ (two-sided paired bootstrap over source groups; 10,000 resamples).}
\label{tab:main-results}
\end{table*}

\section{Experimental Setup}
\label{sec:setup}

\textbf{Models and Evaluation Setup.} We select six models from the Qwen3 and Gemma-3 families for experiments, including Qwen3-8B, Qwen3-14B, Qwen3-32B, and Gemma-3-4B-it, Gemma-3-12B-it, Gemma-3-27B-it, covering parameter scales from 4B to 32B \citep{yang2025qwen3,gemma2025}. All models are evaluated in a zero-shot setting using the same French task instructions and candidate options for original, low, and high inputs. We compute the average token log-probability for each valid answer candidate and apply softmax across the current question's candidates.

\textbf{Evaluation Protocol and Metrics.} Task performance and prediction reliability are evaluated on the 3,360 test triplets and reported separately for original, low, and high inputs. We use accuracy to measure answer correctness, Brier score and NLL to assess the quality of predicted probabilities \citep{gneiting-raftery-2007-proper}, and expected calibration error (ECE) \citep{guo2017} to measure calibration over 10 equal-width confidence bins. For each input condition, accuracy, Brier score, and NLL are averaged over test triplets, giving equal weight to the equally sized dataset–perturbation factor groups. Overall ECE is calculated from all predictions for that condition. For output calibration, we use temperature scaling \citep{guo2017}, fitting a temperature for each model on the original inputs of the held-out set by minimizing NLL. We apply this temperature to the perturbed test inputs to assess calibration transfer. Input restoration uses LanguageTool 6.6\footnote{\url{https://languagetool.org/}} suggestions, accepting edits that follow the rules for apostrophe, diacritic, and keyboard-typo restoration, with the same rules across models and conditions. For perturbation comparisons, metric changes are computed as variant minus original using paired results from the same original question. For the main-table comparisons, we use paired bootstrap significance tests \citep{berg-kirkpatrick-etal-2012-empirical}, resampling source groups to account for triplets that share a source. We also report paired bootstrap confidence intervals (CIs) for perturbation and intervention effects. Prompts, metric definitions, restoration settings, and paired bootstrap procedures, along with implementation and hardware information, are provided in Appendix~\ref{app:evaluation-details}.

\begin{figure*}[t]
\centering
\includegraphics[width=0.90\textwidth,trim=0 0 0 2bp,clip]{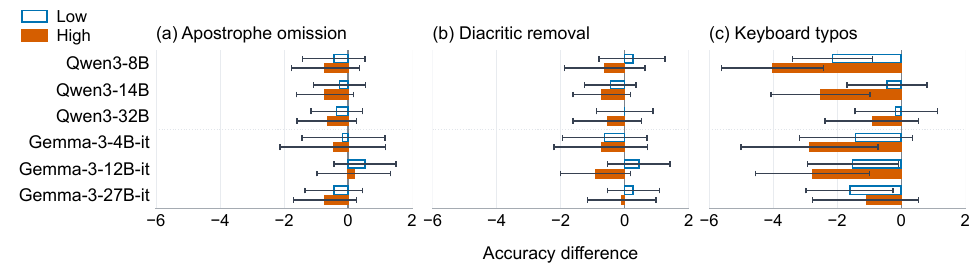}
\caption{Accuracy differences ($\Delta\mathrm{Acc.}=\mathrm{Acc.}_{\mathrm{Variant}}-\mathrm{Acc.}_{\mathrm{Original}}$) under the three perturbation types at low and high intensity. Bars show differences in percentage points, using 160 test triplets from each of seven datasets (1,120 triplets) per perturbation type. Error bars show pointwise 95\% percentile CIs from 10,000 paired bootstrap resamples over source groups. Outlined blue bars represent low intensity, and solid orange bars represent high intensity.}
\label{fig:perturbation-accuracy}
\end{figure*}

\begin{figure*}[t]
\centering
\includegraphics[width=\textwidth,trim=0 0 0 4bp,clip]{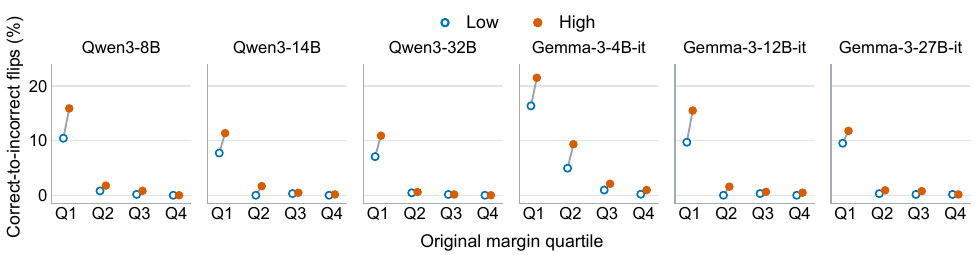}
\caption{Prediction flips under low and high perturbation intensity. For each model, samples predicted correctly under the original input are grouped by the probability margin between the top two candidate answers, from Q1 (smallest margins) to Q4 (largest). Connected points show the proportion of originally correct predictions that become incorrect under each intensity. Q1 concentration intervals are reported in Appendix~\ref{app:confidence-details}.}
\label{fig:margin-flips}
\end{figure*}

\section{Results and Discussion}
\label{sec:results}

\subsection{Task Performance and Prediction Reliability}
\label{sec:performance-probability}
\textbf{Surface perturbations reduce accuracy and increase Brier score and NLL.}
Table~\ref{tab:main-results} presents the overall performance across six models on original inputs and their low- and high-intensity variants. 
Across all models, accuracy consistently decreases as perturbation intensity increases, with high-intensity losses ranging from 0.63 to 1.79 percentage points. Both Brier score and NLL increase monotonically across all six models, indicating deterioration in the quality of predicted probabilities. In contrast, ECE increases relative to original inputs in 9 of the 12 model--intensity conditions, but does not exhibit a consistent trend with intensity.

\textbf{The benefits of model size differ between accuracy and prediction reliability.}
Table~\ref{tab:main-results} further reveals differences between accuracy and prediction reliability across model sizes. Within both model families, larger models exhibit smaller accuracy declines under high-intensity perturbations, but their increases in Brier score and NLL do not consistently decrease with scale. For example, Qwen3-32B shows a smaller accuracy loss than Qwen3-14B but a larger NLL increase, with a similar pattern observed between Gemma-3-27B-it and Gemma-3-12B-it. This indicates that greater resistance to accuracy degradation does not necessarily imply greater stability in predicted probabilities. The effects also vary across tasks: Qwen3-14B shows the largest accuracy decline in commonsense continuation, whereas Gemma-3-12B-it shows the largest decline in NLI. Further task-level comparisons are provided in Appendix~\ref{app:perturbation-results}. 

\Needspace{5\baselineskip}
\textbf{Keyboard typos produce the largest performance degradation at the predefined high-intensity settings.}
Figure~\ref{fig:perturbation-accuracy} compares accuracy changes across the three perturbation types and two intensity levels. 
At the predefined high-intensity settings, keyboard typos produce the largest accuracy drops across all six models, ranging from 0.89 to 4.02 percentage points. They also produce the largest increases in Brier score and NLL, whereas apostrophe omission and diacritic removal have smaller effects. This pattern may partly reflect the effect of character substitutions on word recognition. However, increasing perturbation intensity does not always lead to larger accuracy losses for each model. For example, Gemma-3-27B-it shows a smaller accuracy loss under high- than low-intensity keyboard typos, even as its NLL increases.

\begin{figure*}[!t]
\centering
\includegraphics[width=\textwidth,trim=0 0 0 4bp,clip]{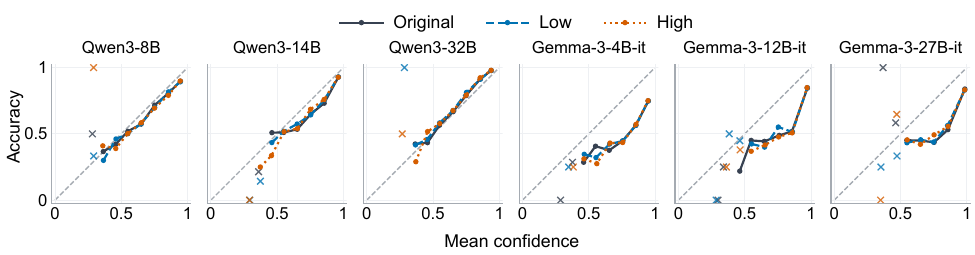}
\caption{Reliability curves for Original, Low, and High inputs before calibration. Each curve compares mean confidence with actual accuracy in 10 equal-width confidence bins, using predictions on the 3,360 test triplets for each condition. The diagonal represents perfect calibration. Crosses show bins with fewer than 30 predictions, and lines connect adjacent bins with at least 30 predictions.}
\label{fig:reliability-curves}
\end{figure*}

\textbf{Takeaway.}
French surface variants consistently reduce accuracy and increase Brier score and NLL. Larger models show smaller accuracy declines but not necessarily smaller increases in Brier score and NLL. The effects also vary across tasks and perturbation types, with keyboard typos causing the largest performance degradation at the evaluated high-intensity levels.

\subsection{Prediction Flips and Confidence Structure}
\label{sec:flips-confidence}

\textbf{Correct-to-incorrect flips are concentrated in samples with small original margins.}
Figure~\ref{fig:margin-flips} shows correct-to-incorrect flip rates across quartiles of the original top-two probability margin, defined as the probability difference between the two most likely candidate answers. Across all six models, the lowest-margin quartile (Q1) exhibits the highest flip rates under both perturbation intensities, with rates increasing consistently from low to high intensity. Flip rates are substantially lower in the higher-margin quartiles, although Gemma-3-4B-it also shows a noticeable effect in Q2. These results reveal a concentration of prediction instability among low-margin samples, despite the relatively modest overall accuracy declines.

\Needspace{4\baselineskip}
\textbf{Confidence--accuracy relationships can shift locally.}
Figure~\ref{fig:reliability-curves} complements the prediction flip analysis by comparing confidence--accuracy relationships across original inputs and their low- and high-intensity variants. The curves remain close over the 0.8--1.0 confidence range, while some lower-confidence bins show larger differences across input conditions. For example, Qwen3-14B's predictions in the 0.4--0.5 confidence bin have mean confidence near 46\% under both original and high-intensity inputs, but their accuracies are 50.91\% and 33.58\%, respectively. This illustrates that similar confidence levels can correspond to different empirical accuracies across input conditions, even when overall ECE changes only slightly (Table~\ref{tab:main-results}).

\Needspace{4\baselineskip}
\textbf{Calibration temperatures remain similar across original and perturbed inputs.}
Despite the local differences in reliability curves, temperatures fitted separately on original, low-intensity, and high-intensity held-out inputs remain similar across conditions (Appendix~\ref{app:confidence-details}). These observations motivate the calibration transfer experiments in the following section.

\textbf{Takeaway.}
Correct-to-incorrect flips are highly concentrated in predictions with small top-two probability margins before perturbation. Surface perturbations can also alter confidence--accuracy relationships within specific confidence ranges, while calibration temperatures remain similar across original and perturbed inputs.

\begin{table*}[!t]
\centering\small\setlength{\tabcolsep}{0.5pt}
\renewcommand{\arraystretch}{1.00}
\begin{tabularx}{\textwidth}{@{}l@{\hspace{4pt}}l@{\hspace{4pt}}*{4}{>{\centering\arraybackslash}X}@{\hspace{4pt}}*{4}{>{\centering\arraybackslash}X}@{\hspace{4pt}}*{4}{>{\centering\arraybackslash}X}@{\hspace{4pt}}*{4}{>{\centering\arraybackslash}X}@{}}
\toprule
 & & \multicolumn{4}{c}{Raw} & \multicolumn{4}{c}{Calibration} & \multicolumn{4}{c}{Restoration} & \multicolumn{4}{c}{Combined} \\
\cmidrule(lr){3-6}\cmidrule(lr){7-10}\cmidrule(lr){11-14}\cmidrule(lr){15-18}
Size & Intensity & Acc. & ECE & Brier & NLL & Acc. & ECE & Brier & NLL & Acc. & ECE & Brier & NLL & Acc. & ECE & Brier & NLL \\
\midrule
\multicolumn{18}{l}{Qwen3} \\
8B & Low & 73.18 & 4.88 & 0.379 & 0.668 & 73.18 & 2.40 & 0.376 & 0.663 & \textbf{73.33} & 5.05 & 0.374 & 0.661 & \textbf{73.33} & \textbf{1.95} & \textbf{0.371} & \textbf{0.655} \\
 & High & 72.17 & 5.61 & 0.388 & 0.681 & 72.17 & 2.43 & 0.384 & 0.675 & \textbf{73.10} & 5.02 & 0.378 & 0.666 & \textbf{73.10} & \textbf{1.85} & \textbf{0.374} & \textbf{0.660} \\
14B & Low & 78.01 & 5.82 & 0.318 & 0.579 & 78.01 & \textbf{2.07} & 0.311 & 0.565 & \textbf{78.21} & 6.15 & 0.316 & 0.575 & \textbf{78.21} & 2.68 & \textbf{0.309} & \textbf{0.560} \\
 & High & 77.08 & 6.02 & 0.323 & 0.589 & 77.08 & 2.57 & 0.317 & 0.576 & \textbf{77.86} & 6.04 & 0.319 & 0.581 & \textbf{77.86} & \textbf{2.23} & \textbf{0.312} & \textbf{0.566} \\
32B & Low & 80.48 & 4.20 & 0.283 & 0.520 & 80.48 & 1.93 & 0.281 & 0.508 & \textbf{80.65} & 4.07 & 0.278 & 0.511 & \textbf{80.65} & \textbf{1.03} & \textbf{0.276} & \textbf{0.499} \\
 & High & 79.97 & 4.57 & 0.289 & 0.529 & 79.97 & \textbf{1.50} & 0.287 & 0.517 & \textbf{80.30} & 3.97 & 0.281 & 0.515 & \textbf{80.30} & 1.56 & \textbf{0.279} & \textbf{0.503} \\
\midrule
\multicolumn{18}{l}{Gemma-3 (instruction-tuned)} \\
4B & Low & 61.67 & 24.09 & 0.603 & 1.127 & 61.67 & \textbf{5.66} & 0.512 & 0.921 & \textbf{61.70} & 24.55 & 0.606 & 1.132 & \textbf{61.70} & 6.11 & \textbf{0.511} & \textbf{0.918} \\
 & High & 61.07 & 24.19 & 0.608 & 1.130 & 61.07 & \textbf{5.55} & 0.516 & 0.927 & \textbf{61.70} & 24.32 & 0.607 & 1.132 & \textbf{61.70} & 6.00 & \textbf{0.512} & \textbf{0.920} \\
12B & Low & \textbf{75.68} & 16.17 & 0.399 & 0.796 & \textbf{75.68} & \textbf{5.65} & 0.345 & 0.634 & 75.54 & 16.59 & 0.398 & 0.795 & 75.54 & 6.03 & \textbf{0.343} & \textbf{0.628} \\
 & High & 74.70 & 16.76 & 0.407 & 0.806 & 74.70 & \textbf{6.07} & 0.351 & 0.645 & \textbf{74.91} & 17.04 & 0.403 & 0.801 & \textbf{74.91} & 6.24 & \textbf{0.346} & \textbf{0.634} \\
27B & Low & 78.07 & 16.69 & 0.377 & 0.891 & 78.07 & 3.74 & 0.319 & 0.598 & \textbf{78.54} & 16.24 & 0.369 & 0.873 & \textbf{78.54} & \textbf{3.61} & \textbf{0.313} & \textbf{0.587} \\
 & High & 78.04 & 16.63 & 0.379 & 0.897 & 78.04 & \textbf{3.10} & 0.323 & 0.605 & \textbf{78.27} & 16.41 & 0.371 & 0.879 & \textbf{78.27} & 3.41 & \textbf{0.316} & \textbf{0.592} \\
\bottomrule
\end{tabularx}
\caption{Four-method comparison on 3,360 test triplets, reported separately for each model and perturbation intensity. Acc. and ECE are percentages; Brier and NLL retain their original scales. Combined applies the original-input temperature after restoration. Bold marks the highest accuracy and lowest ECE, Brier and NLL within each row. Additional intervention results and paired bootstrap CIs are provided in Appendices~\ref{app:interventions} and~\ref{app:effect-intervals}, respectively.}
\label{tab:intervention-summary}
\end{table*}

\Needspace{8\baselineskip}
\subsection{Calibration Transfer and Input Restoration}
\label{sec:calibration-restoration}

\textbf{Temperature calibration on original inputs transfers to both perturbation intensities.}
Building on the temperature stability observed in \S\ref{sec:flips-confidence}, we evaluate whether calibration fitted on original inputs transfers to their surface variants. Table~\ref{tab:intervention-summary} compares four intervention settings across six models and two perturbation intensities. We first examine the Raw and Calibration results to assess calibration transfer. Applying the temperature fitted on original held-out inputs reduces ECE, Brier score, and NLL for all six models at both intensities, while leaving accuracy unchanged. The magnitude of these improvements varies across models, with particularly large ECE and Brier score reductions for the Gemma-3 models.

\textbf{Input restoration reduces accuracy losses under high-intensity perturbations.}
Unlike temperature calibration, which adjusts predicted probabilities, input restoration attempts to correct perturbations in the input text. Table~\ref{tab:intervention-summary} also compares the Raw and Restoration results across six models. Input restoration improves accuracy for five of the six models at low intensity and for all six models at high intensity. Specifically, the high-intensity gains range from 0.21 to 0.92 percentage points. These results indicate that input restoration can mitigate accuracy degradation under both perturbation intensities, with more consistent and generally larger improvements at high intensity.

\textbf{Combining restoration and calibration preserves restored accuracy and further lowers Brier score and NLL.}
Table~\ref{tab:intervention-summary} further shows that applying the original-input temperature after input restoration preserves the accuracy while achieving lower Brier score and NLL than either intervention alone across all 12 model--intensity conditions. Although ECE does not consistently favor the combined approach, these findings highlight the complementary benefits of input restoration and output calibration, providing a practical reference for jointly addressing accuracy and reliability under French surface variations.

\textbf{Takeaway.}
For French surface variants, calibration fitted on original inputs reduces ECE, Brier score and NLL at both perturbation intensities, while input restoration generally improves accuracy, especially at high perturbation intensity. Their combination preserves restored accuracy and achieves lower Brier score and NLL than either method alone.

\section{Conclusion}
\label{sec:conclusion}
This paper presents a French benchmark covering seven task types, with each original input paired with low- and high-intensity variants. The results show that increasing perturbation intensity consistently reduces accuracy and increases Brier score and NLL across six models. Larger models are more robust in accuracy under high intensity, while their Brier score and NLL do not show consistent benefits from scale. Keyboard typos lead to greater degradation than the other variations at the predefined high-intensity settings. Calibration transfers from original inputs to variants and complements restoration, retaining its accuracy gains while improving Brier score and NLL beyond either method alone. The benchmark provides a basis for further research on accuracy and prediction reliability under French writing variations, with potential extensions to multilingual evaluation.

\section*{Limitations}
\label{sec:limitations}
This paper focuses on three types of surface variations in French, applied separately at two predefined perturbation intensities. Real user inputs may combine these variations in proportions that differ across users and settings. The benchmark evaluates each perturbation type separately, leaving their joint effects untested.

Our benchmark covers seven task types drawn from existing French datasets. Although these datasets include diverse texts, they do not cover all contexts of French language use. Future work could expand the benchmark to include additional domains and text types.

\section*{Ethical Considerations}
\label{sec:ethics}
This benchmark is constructed from existing datasets, following their licenses and redistribution requirements. The research team did not have access to the annotators’ personal identities. The benchmark is intended for research on model performance and prediction reliability under variations in French input.

The writing variations studied here should not be used to infer users' language proficiency or social identity. We encourage the use of this benchmark to improve model support for diverse writing practices.

\section*{Acknowledgments}
We used ChatGPT and Codex for literature search, manuscript drafting and language editing, translation, and code development and debugging assistance. The authors reviewed the manuscript and take responsibility for its content and results.

\bibliography{main}

\clearpage
\appendix
\section{Benchmark Construction and Quality Review}
\label{app:construction}

Each triplet contains one original input and its low- and high-intensity variants. The benchmark contains 4,200 triplets (12,600 inputs), split into 840 held-out and 3,360 test triplets.

\subsection{Perturbation Rules}
\label{app:perturbation-rules}
Apostrophe omission removes the apostrophe from eligible French elisions while retaining the surrounding letters. Highly lexicalized forms such as \textit{aujourd'hui} and predefined lexical collisions such as \textit{d'eux} $\rightarrow$ \textit{deux} are excluded. Diacritic removal replaces accented letters with their unaccented forms. Predefined ambiguous forms, including \textit{à}, \textit{où}, \textit{dès}, \textit{sûr}, and \textit{dû}, are excluded. Other word-form overlaps arising from apostrophe omission or diacritic removal, such as \textit{d'une} $\rightarrow$ \textit{dune} and \textit{divisé} $\rightarrow$ \textit{divise}, remain eligible and are assessed in context during quality review. Keyboard typos replace one non-initial letter in an eligible word of at least four letters with an AZERTY-neighbor letter. Neighbors are horizontally, vertically, or diagonally adjacent positions in the three letter rows of the AZERTY layout. Candidate substitutions that form another common French word are excluded using wordfreq 3.1.1\footnote{\url{https://github.com/rspeer/wordfreq}} with a Zipf-frequency threshold of 2.5.

Task headers, structural markers, numerical values, LaTeX commands, and marked MCL-WiC target words are protected. For mathematical questions, the answer-option block is protected under diacritic removal and keyboard typos. Apostrophe omission may also be applied to eligible French elisions in the option text. Answer codes, reference labels, and option order are shared within each original--low--high triplet. Both variants use the same edit plan, with all low-intensity edits retained at high intensity.

\label{app:example-glosses}
The original words in Figure~\ref{fig:construction} mean: \textit{d'un} (``of a''), \textit{l'étui} (``the case''), \textit{divisé} (``divided''), \textit{résultant} (``resulting''), \textit{tous} (``all''), \textit{cosets} (``cosets''), and \textit{sous-groupe} (``subgroup'').

\subsection{Automatic Task-Impact Review}
\label{app:automatic-review}
DeepSeek-V4-Pro and Gemini-2.5-Pro independently screen all candidate triplets using a shared instruction, five synthetic examples, and the same decision rules. Each variant receives four structured assessments and a brief explanation. The core instruction is summarized below.

\begin{quote}\small
Compare each low- and high-intensity variant with the original input. Check whether the edits introduce a task-relevant difference, conflict with the reference answer, exceed the specified surface operation, or damage the required task structure. Assess the original question first and record source-question issues separately. Evaluate the two variants independently. Recoverable spelling changes and reduced fluency alone do not constitute failure. Return a structured assessment and a short reason for each variant.
\end{quote}

A variant is assigned \textit{fail} if any assessment identifies a clear task-relevant difference, an incompatible reference answer, an edit outside the specified operation, or broken task structure; otherwise, a valid assessment maps to \textit{pass}. A variant is assigned \textit{pass} when a valid assessment does not provide clear evidence that a predefined failure criterion is met. A triplet is flagged when either variant fails. Missing or invalid responses are recorded separately. The research team reviewed the DeepSeek flag and judged it to be a false positive. Triplets with failed or invalid Gemini assessments underwent targeted human review. Replacement triplets were accepted after both models returned valid \textit{pass} judgments for both variants.

\subsection{Human Review Criteria and Outcomes}
\label{app:human-review}
Reviewers compare each low- and high-intensity variant with its original input and reference answer. The two variants are judged separately. A variant receives \textit{fail} when a clear task-relevant change, direct incompatibility with the reference answer, an unreadable passage, loss of required task structure, or an out-of-scope edit is observed; otherwise, it receives \textit{pass}. Source-task defects that prevent answering, clearly incorrect reference answers, or multiple clearly correct options also lead to \textit{fail} judgments.

Out-of-scope edits include adding, deleting, or moving whole words; deleting letters or words that express negation; changing numerical values; and replacing expressions in ways that cannot result from the three specified surface operations. Reviewers check sentiment orientation for review ratings, the premise--hypothesis relation for NLI, and the evidence needed to choose among options for reading, mathematics, continuation, and moral-action tasks. For MCL-WiC, both marked target words must remain intact. Recoverable spelling changes in answer-option text are judged by the same criteria as changes in the question.

The random audit covered 210 triplets, whose 420 low- and high-intensity variants were rated separately by three reviewers. Initial Krippendorff's $\alpha$ and unanimous agreement were 0.969 and 98.33\%, respectively. Following clarification of the annotation guidelines, particularly the distinction between recoverable spelling changes and task-relevant changes, the final accepted ratings reached 1.000 and 100.00\%, respectively.

Three triplets were replaced following majority judgments, and one additional triplet was replaced after a source-question check. The four replacements retained their dataset, perturbation factor, and split assignments and passed the two-model task-impact review.

\section{Experimental Details}
\label{app:inference}
\label{app:evaluation-details}

\subsection{Implementation and Hardware}
\label{app:implementation}
The model checkpoints are Qwen/Qwen3-8B, Qwen/Qwen3-14B, Qwen/Qwen3-32B, google/gemma-3-4b-it, google/gemma-3-12b-it, and google/gemma-3-27b-it. Each inference run uses one GPU, with tensor parallel size set to one. Qwen3-8B, Qwen3-14B, Gemma-3-4B-it, and Gemma-3-12B-it were evaluated on an NVIDIA RTX 5090 with 32\,GB of memory. Qwen3-32B and Gemma-3-27B-it were evaluated on an NVIDIA RTX PRO 6000 Blackwell Server Edition with 96\,GB. Both environments use vLLM 0.19.1, PyTorch 2.10.0 with CUDA 12.8 (\texttt{cu128}), and Transformers 5.6.2.

Automatic quality screening used the API model identifiers \texttt{deepseek-v4-pro} and \texttt{gemini-2.5-pro}. Requests used low reasoning effort and a maximum output length of 4,096 tokens. The screening protocol is described in Appendix~\ref{app:automatic-review}.

\subsection{Prompt Template}
\label{app:prompts}
All tasks use the following French scoring prefix. The task-specific instruction is given in Table~\ref{tab:task-instructions}; the input and answer options are filled from the corresponding benchmark condition. Qwen3 uses the literal header \texttt{[Système]}, while Gemma-3 uses \texttt{[Instruction]}. Both use the same instruction text. Options are rendered once, in answer-code order, and only the valid options for the task are included.

\begin{quote}\small
\texttt{[Système]}\\
Vous répondez à une tâche en français à choix multiple. Choisissez exactement une option parmi celles qui sont fournies.\\[5pt]
\texttt{[Tâche]}\\
\textit{Task-specific instruction}\\[5pt]
\texttt{[Entrée]}\\
\textit{Task input}\\[5pt]
\texttt{[Options]}\\
\texttt{(A)} \textit{Option A}\\
\texttt{(B)} \textit{Option B}\\
\textit{Remaining valid options, when applicable}\\[5pt]
\texttt{[Réponse]}\\
\texttt{La bonne option est (}
\end{quote}

In English, the common instruction asks the model to answer a French multiple-choice task and select exactly one of the provided options. The headers denote System (or Instruction), Task, Input, Options, and Response. The answer prefix means ``The correct option is (''.

\begin{table*}[t]
\centering\small
\begin{tabularx}{\textwidth}{@{}p{0.19\textwidth}X@{}}
\toprule
Task & French instruction \\
\midrule
Review rating & Déterminez le degré de satisfaction envers le produit exprimé dans cet avis client Amazon. \\
NLI & Déterminez si l'hypothèse découle de la prémisse, la contredit, ou si leur relation reste indéterminée. \\
Reading comprehension & Répondez à la question en vous fondant uniquement sur le passage. \\
Mathematics & Résolvez cette question de mathématiques. \\
Continuation & Identifiez la continuation la plus plausible et la plus cohérente avec le contexte. \\
Moral action & Identifiez l'action qui respecte le mieux les normes morales et sociales ordinaires compte tenu de la situation et de l'intention. \\
Word in context & Déterminez si les deux mots placés entre $[\![$ $]\!]$ sont employés dans le même sens dans les deux phrases. \\
\bottomrule
\end{tabularx}
\caption{Task-specific instructions. The double brackets in the word-in-context instruction mark the two target occurrences.}
\label{tab:task-instructions}
\end{table*}

The task instructions translate as follows: determine the satisfaction level expressed in the Amazon review; determine whether the hypothesis follows from, contradicts, or is undetermined by the premise; answer using only the passage; solve the mathematics question; choose the most plausible and coherent continuation; choose the action that best respects ordinary moral and social norms given the situation and intention; and determine whether the two bracketed words have the same meaning in their respective sentences.

\subsection{Candidate Scoring and Runtime}
Each valid answer code is scored as a continuation of the prefix, for example \texttt{A)} or \texttt{B)}. We average the token log probabilities within each continuation and apply softmax across the candidate scores. The highest-probability candidate determines the prediction and its confidence. The evaluation uses these candidate scores, with no chain-of-thought generation. All six models use bfloat16 (BF16) precision without quantization, a maximum context length of 4,096 tokens, and a row batch size of eight in vLLM 0.19.1. Scoring requests use \texttt{prompt\_logprobs=64}, \texttt{temperature=0}, and \texttt{max\_tokens=1}. Candidate scores are obtained from prompt-token log probabilities; the additional generated token is not used for prediction. Original, perturbed, and restored inputs use the same scoring procedure.

\subsection{Input Restoration}
\label{app:restoration-method}
We use a local LanguageTool 6.6 instance with French language settings. For each reported match, the first suggested replacement is considered. Suggestions identified as spelling or typographic edits are retained when they restore a French apostrophe, restore diacritics without changing the unaccented letter sequence, or replace a single non-initial letter with an AZERTY neighbor. All three restoration operations are available for every input, with the same rules across models and conditions. Non-overlapping accepted edits are applied to the input before scoring. The procedure is applied separately to the original, low-intensity, and high-intensity inputs of every test triplet; the original inputs provide the control reported in Table~\ref{tab:restoration-full}.

\subsection{Temperature Fitting and Metrics}
Let $s_y(x)$ denote the mean token log probability of candidate answer code $y$ for input $x$, and let $\mathcal{Y}(x)$ be its candidate set. Temperature scaling gives
\[
p_T(y\mid x)=\frac{\exp(s_y(x)/T)}{\sum_{k\in\mathcal{Y}(x)}\exp(s_k(x)/T)}.
\]
The uncalibrated probabilities use $T=1$.

Each model's temperature is fitted by minimizing NLL on the original inputs of the 840 held-out triplets, then fixed for evaluation on original, low-intensity, high-intensity, and restored test inputs. We search $T\in[0.05,200]$ on a 201-point logarithmic grid, then refine around the best point with 80 ternary-search iterations. Condition-specific temperatures are fitted separately on the corresponding low- or high-intensity variants of the same held-out triplets. The combined method applies the original temperature to the scores obtained after input restoration.

Accuracy is the proportion of correct predictions. Brier score is the mean sum of squared differences between the candidate probabilities and the one-hot reference labels. NLL is the mean negative log probability assigned to the reference answer. ECE uses ten equal-width confidence bins and weights each bin's absolute confidence--accuracy difference by its sample count. Overall ECE is computed from pooled predictions. The top-two probability margin, defined as the highest candidate probability minus the second highest, is used for the prediction-flip analysis; the deletion control uses the correct-candidate log-score margin defined in Appendix~\ref{app:confidence-details}.

\subsection{Paired Statistical Comparisons}
\label{app:statistics}
For each model, perturbation comparisons match original predictions with low- or high-intensity predictions within the same test triplets. Bootstrap units are shared passages for Belebele, groups connected by shared lemmas or source sentences for MCL-WiC, and individual questions for the other datasets. We draw 10,000 paired resamples of source groups, retaining every triplet in each sampled group, and recompute the metric differences. Two-sided tests compare the absolute observed difference with absolute centered bootstrap differences, using an add-one correction. Holm adjustment is applied jointly across six models, two intensities, and four metrics; stars mark adjusted $p<0.05$. CIs are pointwise 95\% percentile intervals without multiplicity adjustment.

Intervention comparisons use the same resampling procedure, including recomputation of pooled ECE. Temperatures remain fixed at their original held-out estimates, so the intervals describe test-sample uncertainty for these fitted temperatures.

\FloatBarrier

\FloatBarrier
\section{Detailed Perturbation Results}\label{app:perturbation-results}

The following tables give the dataset- and factor-level results for Section~\ref{sec:performance-probability}, using the test triplets from our internal split. Each dataset--factor group contains 160 triplets, evaluated separately under Original, Low, and High. Acc. and ECE are reported as percentages; Brier score and NLL are reported on their original scales. Apostrophe, Diacritic, and Keyboard denote apostrophe omission, diacritic removal, and keyboard typos, respectively.

MMLU-Math denotes the French mathematics subset of Global-MMLU; Histoires denotes HistoiresMorales.

\begin{table*}[tp]
\centering\small\setlength{\tabcolsep}{3pt}
\renewcommand{\arraystretch}{1.08}
\begin{tabular*}{\textwidth}{@{\extracolsep{\fill}}llcccc@{}}
\toprule
Dataset & Factor & Acc. & ECE & Brier & NLL \\
\midrule
\multicolumn{6}{l}{Qwen3-8B} \\
MARC & Apostrophe & 55.63 / 57.50 / 57.50 & 17.39 / 16.47 / 17.37 & 0.566 / 0.567 / 0.563 & 0.983 / 0.990 / 0.988 \\
 & Diacritic & 55.63 / 55.00 / 53.75 & 18.52 / 19.57 / 19.37 & 0.617 / 0.623 / 0.624 & 1.110 / 1.113 / 1.111 \\
 & Keyboard & 56.88 / 56.25 / 50.63 & 16.16 / 15.48 / 20.46 & 0.596 / 0.596 / 0.632 & 1.057 / 1.064 / 1.120 \\
Belebele & Apostrophe & 93.75 / 93.13 / 94.38 & 6.66 / 6.78 / 5.44 & 0.102 / 0.101 / 0.104 & 0.221 / 0.222 / 0.229 \\
 & Diacritic & 87.50 / 90.63 / 89.38 & 3.12 / 2.65 / 1.49 & 0.171 / 0.155 / 0.168 & 0.347 / 0.322 / 0.347 \\
 & Keyboard & 92.50 / 91.25 / 91.25 & 3.85 / 6.40 / 5.57 & 0.110 / 0.123 / 0.129 & 0.223 / 0.246 / 0.266 \\
FrenchNLI & Apostrophe & 82.50 / 82.50 / 81.25 & 5.63 / 5.67 / 4.38 & 0.269 / 0.275 / 0.290 & 0.499 / 0.506 / 0.528 \\
 & Diacritic & 80.63 / 81.88 / 81.25 & 4.55 / 6.24 / 5.35 & 0.285 / 0.289 / 0.287 & 0.514 / 0.523 / 0.521 \\
 & Keyboard & 75.63 / 71.88 / 67.50 & 7.71 / 6.02 / 6.69 & 0.342 / 0.380 / 0.403 & 0.587 / 0.644 / 0.681 \\
MMLU-Math & Apostrophe & 65.00 / 65.00 / 65.00 & 9.93 / 10.88 / 10.52 & 0.445 / 0.444 / 0.450 & 0.836 / 0.843 / 0.844 \\
 & Diacritic & 63.13 / 63.75 / 61.25 & 8.40 / 9.55 / 10.85 & 0.493 / 0.495 / 0.517 & 0.906 / 0.909 / 0.952 \\
 & Keyboard & 66.88 / 65.63 / 63.75 & 9.98 / 8.87 / 8.96 & 0.469 / 0.471 / 0.490 & 0.881 / 0.897 / 0.916 \\
HellaSwag & Apostrophe & 78.75 / 75.63 / 75.00 & 10.19 / 8.65 / 6.79 & 0.352 / 0.363 / 0.363 & 0.686 / 0.704 / 0.704 \\
 & Diacritic & 74.38 / 74.38 / 75.63 & 12.20 / 11.55 / 11.79 & 0.342 / 0.351 / 0.346 & 0.664 / 0.677 / 0.670 \\
 & Keyboard & 78.13 / 76.25 / 75.00 & 12.18 / 15.72 / 12.06 & 0.334 / 0.355 / 0.368 & 0.670 / 0.707 / 0.727 \\
Histoires & Apostrophe & 89.38 / 88.75 / 87.50 & 8.30 / 7.55 / 6.15 & 0.166 / 0.179 / 0.180 & 0.293 / 0.307 / 0.307 \\
 & Diacritic & 90.63 / 90.00 / 89.38 & 9.62 / 10.25 / 8.91 & 0.161 / 0.168 / 0.172 & 0.291 / 0.299 / 0.304 \\
 & Keyboard & 90.00 / 86.88 / 87.50 & 7.70 / 8.32 / 8.68 & 0.161 / 0.180 / 0.185 & 0.284 / 0.310 / 0.318 \\
MCL-WiC & Apostrophe & 58.13 / 57.50 / 57.50 & 28.06 / 28.43 / 28.61 & 0.605 / 0.615 / 0.621 & 0.910 / 0.920 / 0.925 \\
 & Diacritic & 60.63 / 58.75 / 57.50 & 25.10 / 27.14 / 28.44 & 0.560 / 0.575 / 0.577 & 0.852 / 0.862 / 0.863 \\
 & Keyboard & 57.50 / 54.38 / 53.75 & 28.79 / 32.47 / 32.67 & 0.617 / 0.658 / 0.678 & 0.927 / 0.971 / 0.988 \\
\midrule
\multicolumn{6}{l}{Qwen3-14B} \\
MARC & Apostrophe & 48.75 / 50.63 / 50.00 & 29.83 / 30.30 / 29.78 & 0.736 / 0.729 / 0.727 & 1.384 / 1.374 / 1.364 \\
 & Diacritic & 46.88 / 46.25 / 45.00 & 31.17 / 31.15 / 33.04 & 0.762 / 0.767 / 0.769 & 1.412 / 1.424 / 1.434 \\
 & Keyboard & 51.88 / 50.00 / 48.75 & 25.20 / 26.58 / 27.59 & 0.709 / 0.742 / 0.758 & 1.343 / 1.405 / 1.441 \\
Belebele & Apostrophe & 95.00 / 95.00 / 93.75 & 5.08 / 4.92 / 4.24 & 0.078 / 0.074 / 0.085 & 0.161 / 0.156 / 0.175 \\
 & Diacritic & 92.50 / 92.50 / 91.88 & 4.03 / 3.52 / 2.06 & 0.121 / 0.125 / 0.138 & 0.245 / 0.253 / 0.294 \\
 & Keyboard & 93.75 / 93.75 / 92.50 & 4.22 / 6.03 / 6.13 & 0.092 / 0.090 / 0.100 & 0.188 / 0.192 / 0.217 \\
FrenchNLI & Apostrophe & 80.00 / 80.63 / 79.38 & 8.34 / 6.67 / 7.58 & 0.301 / 0.298 / 0.299 & 0.547 / 0.546 / 0.559 \\
 & Diacritic & 78.13 / 78.75 / 79.38 & 7.36 / 9.64 / 9.24 & 0.338 / 0.326 / 0.323 & 0.606 / 0.586 / 0.583 \\
 & Keyboard & 73.13 / 70.63 / 70.00 & 15.22 / 14.13 / 10.68 & 0.423 / 0.431 / 0.430 & 0.765 / 0.767 / 0.740 \\
MMLU-Math & Apostrophe & 76.88 / 75.00 / 76.25 & 5.69 / 8.09 / 7.66 & 0.341 / 0.345 / 0.344 & 0.678 / 0.683 / 0.681 \\
 & Diacritic & 76.25 / 75.00 / 76.25 & 7.01 / 9.70 / 10.37 & 0.338 / 0.340 / 0.339 & 0.626 / 0.621 / 0.623 \\
 & Keyboard & 70.00 / 71.88 / 70.00 & 8.93 / 6.82 / 7.60 & 0.386 / 0.398 / 0.411 & 0.721 / 0.745 / 0.762 \\
HellaSwag & Apostrophe & 84.38 / 81.25 / 81.25 & 5.11 / 8.05 / 5.40 & 0.238 / 0.244 / 0.251 & 0.469 / 0.474 / 0.488 \\
 & Diacritic & 85.00 / 85.00 / 83.13 & 3.90 / 4.40 / 3.56 & 0.223 / 0.228 / 0.234 & 0.426 / 0.437 / 0.448 \\
 & Keyboard & 85.63 / 83.75 / 80.00 & 5.65 / 8.12 / 8.52 & 0.216 / 0.225 / 0.244 & 0.407 / 0.443 / 0.472 \\
Histoires & Apostrophe & 94.38 / 93.13 / 93.13 & 1.45 / 1.73 / 1.68 & 0.090 / 0.095 / 0.098 & 0.154 / 0.164 / 0.167 \\
 & Diacritic & 93.75 / 93.75 / 94.38 & 2.49 / 1.75 / 1.60 & 0.095 / 0.094 / 0.091 & 0.164 / 0.166 / 0.161 \\
 & Keyboard & 95.00 / 94.38 / 92.50 & 2.78 / 5.17 / 0.68 & 0.089 / 0.097 / 0.114 & 0.158 / 0.166 / 0.193 \\
MCL-WiC & Apostrophe & 70.00 / 71.88 / 70.63 & 9.54 / 8.53 / 9.14 & 0.387 / 0.387 / 0.390 & 0.578 / 0.579 / 0.582 \\
 & Diacritic & 82.50 / 80.63 / 80.00 & 5.22 / 6.51 / 7.13 & 0.261 / 0.263 / 0.271 & 0.422 / 0.426 / 0.432 \\
 & Keyboard & 72.50 / 74.38 / 70.63 & 9.41 / 8.65 / 5.64 & 0.367 / 0.372 / 0.378 & 0.548 / 0.560 / 0.562 \\
\bottomrule
\end{tabular*}
\caption{Uncalibrated results by model, dataset, and perturbation factor. Each cell lists Original / Low / High. Acc. and ECE are percentages; Brier and NLL retain their original scales.}
\label{tab:detailed-perturbation}
\end{table*}

\begin{table*}[tp]
\centering\small\setlength{\tabcolsep}{3pt}
\renewcommand{\arraystretch}{1.08}
\addtocounter{table}{-1}
\renewcommand{\theHtable}{\arabic{table}.part2}
\begin{tabular*}{\textwidth}{@{\extracolsep{\fill}}llcccc@{}}
\toprule
Dataset & Factor & Acc. & ECE & Brier & NLL \\
\midrule
\multicolumn{6}{l}{Qwen3-32B} \\
MARC & Apostrophe & 57.50 / 56.88 / 57.50 & 11.31 / 11.98 / 14.36 & 0.578 / 0.582 / 0.582 & 1.015 / 1.024 / 1.025 \\
 & Diacritic & 51.88 / 51.25 / 51.25 & 14.61 / 15.34 / 14.03 & 0.610 / 0.614 / 0.614 & 1.085 / 1.092 / 1.089 \\
 & Keyboard & 56.25 / 53.75 / 52.50 & 9.90 / 11.12 / 12.47 & 0.548 / 0.568 / 0.605 & 1.005 / 1.041 / 1.101 \\
Belebele & Apostrophe & 94.38 / 95.63 / 94.38 & 10.08 / 11.06 / 10.04 & 0.088 / 0.086 / 0.093 & 0.222 / 0.218 / 0.229 \\
 & Diacritic & 94.38 / 93.75 / 92.50 & 10.07 / 11.95 / 10.71 & 0.119 / 0.129 / 0.132 & 0.280 / 0.298 / 0.309 \\
 & Keyboard & 96.25 / 95.63 / 95.63 & 11.09 / 12.76 / 13.74 & 0.080 / 0.098 / 0.103 & 0.207 / 0.245 / 0.258 \\
FrenchNLI & Apostrophe & 82.50 / 81.88 / 82.50 & 8.19 / 4.72 / 4.08 & 0.282 / 0.277 / 0.277 & 0.510 / 0.507 / 0.506 \\
 & Diacritic & 81.25 / 81.25 / 81.88 & 9.96 / 6.64 / 8.09 & 0.265 / 0.272 / 0.274 & 0.474 / 0.486 / 0.490 \\
 & Keyboard & 74.38 / 75.00 / 73.75 & 6.67 / 7.31 / 7.59 & 0.354 / 0.349 / 0.368 & 0.593 / 0.592 / 0.622 \\
MMLU-Math & Apostrophe & 78.75 / 78.13 / 78.13 & 6.60 / 7.57 / 7.87 & 0.319 / 0.325 / 0.321 & 0.618 / 0.635 / 0.629 \\
 & Diacritic & 74.38 / 76.25 / 75.00 & 10.40 / 10.50 / 9.61 & 0.343 / 0.345 / 0.348 & 0.664 / 0.671 / 0.673 \\
 & Keyboard & 72.50 / 74.38 / 71.88 & 7.70 / 8.15 / 10.21 & 0.370 / 0.379 / 0.394 & 0.701 / 0.722 / 0.750 \\
HellaSwag & Apostrophe & 88.75 / 90.00 / 89.38 & 13.21 / 14.35 / 13.73 & 0.194 / 0.206 / 0.207 & 0.413 / 0.432 / 0.435 \\
 & Diacritic & 86.88 / 85.00 / 85.00 & 11.44 / 10.64 / 9.01 & 0.201 / 0.212 / 0.212 & 0.412 / 0.428 / 0.429 \\
 & Keyboard & 86.88 / 88.13 / 86.25 & 11.10 / 13.61 / 13.64 & 0.187 / 0.216 / 0.234 & 0.388 / 0.456 / 0.478 \\
Histoires & Apostrophe & 91.88 / 92.50 / 92.50 & 4.63 / 6.46 / 8.28 & 0.121 / 0.121 / 0.123 & 0.220 / 0.222 / 0.224 \\
 & Diacritic & 93.75 / 93.75 / 92.50 & 6.71 / 6.75 / 7.73 & 0.101 / 0.103 / 0.104 & 0.196 / 0.198 / 0.198 \\
 & Keyboard & 93.75 / 93.13 / 92.50 & 8.96 / 8.06 / 8.48 & 0.094 / 0.115 / 0.119 & 0.182 / 0.211 / 0.221 \\
MCL-WiC & Apostrophe & 80.63 / 76.88 / 75.63 & 10.60 / 7.38 / 7.41 & 0.344 / 0.351 / 0.351 & 0.519 / 0.525 / 0.526 \\
 & Diacritic & 82.50 / 83.75 / 83.13 & 8.46 / 8.96 / 7.98 & 0.266 / 0.261 / 0.270 & 0.415 / 0.411 / 0.419 \\
 & Keyboard & 74.38 / 73.13 / 75.63 & 7.80 / 10.00 / 7.45 & 0.338 / 0.339 / 0.339 & 0.504 / 0.508 / 0.503 \\
\midrule
\multicolumn{6}{l}{Gemma-3-4B-it} \\
MARC & Apostrophe & 35.63 / 35.63 / 36.88 & 53.30 / 53.54 / 51.88 & 1.099 / 1.099 / 1.078 & 2.252 / 2.248 / 2.211 \\
 & Diacritic & 33.13 / 31.25 / 30.63 & 56.90 / 57.17 / 57.79 & 1.170 / 1.174 / 1.174 & 2.425 / 2.410 / 2.399 \\
 & Keyboard & 38.13 / 36.88 / 35.00 & 50.08 / 51.76 / 53.28 & 1.071 / 1.078 / 1.090 & 2.195 / 2.224 / 2.225 \\
Belebele & Apostrophe & 85.63 / 84.38 / 82.50 & 5.82 / 4.68 / 7.06 & 0.241 / 0.251 / 0.268 & 0.517 / 0.527 / 0.553 \\
 & Diacritic & 85.63 / 85.00 / 85.00 & 5.98 / 5.88 / 7.73 & 0.251 / 0.257 / 0.266 & 0.549 / 0.547 / 0.571 \\
 & Keyboard & 84.38 / 79.38 / 76.88 & 5.71 / 7.89 / 9.78 & 0.269 / 0.324 / 0.374 & 0.575 / 0.662 / 0.745 \\
FrenchNLI & Apostrophe & 72.50 / 71.25 / 72.50 & 14.50 / 15.31 / 14.39 & 0.417 / 0.429 / 0.417 & 0.741 / 0.756 / 0.735 \\
 & Diacritic & 65.00 / 66.25 / 67.50 & 22.31 / 21.00 / 19.19 & 0.550 / 0.538 / 0.508 & 0.983 / 0.947 / 0.900 \\
 & Keyboard & 63.75 / 63.75 / 67.50 & 24.68 / 22.92 / 17.11 & 0.578 / 0.568 / 0.508 & 0.985 / 0.977 / 0.891 \\
MMLU-Math & Apostrophe & 46.25 / 50.00 / 48.13 & 30.32 / 27.99 / 29.06 & 0.770 / 0.771 / 0.771 & 1.499 / 1.515 / 1.488 \\
 & Diacritic & 35.63 / 34.38 / 34.38 & 41.37 / 41.86 / 41.54 & 0.977 / 0.972 / 0.969 & 1.860 / 1.832 / 1.835 \\
 & Keyboard & 38.75 / 39.38 / 35.00 & 37.35 / 34.85 / 38.42 & 0.897 / 0.884 / 0.922 & 1.714 / 1.672 / 1.707 \\
HellaSwag & Apostrophe & 63.75 / 61.88 / 56.25 & 16.87 / 17.32 / 21.99 & 0.562 / 0.569 / 0.623 & 1.091 / 1.104 / 1.181 \\
 & Diacritic & 67.50 / 66.88 / 67.50 & 14.41 / 15.46 / 14.53 & 0.507 / 0.517 / 0.502 & 0.994 / 1.001 / 0.977 \\
 & Keyboard & 56.25 / 56.25 / 51.25 & 24.52 / 21.53 / 25.29 & 0.673 / 0.657 / 0.699 & 1.301 / 1.264 / 1.296 \\
Histoires & Apostrophe & 87.50 / 88.75 / 88.75 & 8.17 / 6.69 / 9.00 & 0.211 / 0.195 / 0.199 & 0.375 / 0.356 / 0.363 \\
 & Diacritic & 92.50 / 91.88 / 88.75 & 3.75 / 4.65 / 6.82 & 0.136 / 0.146 / 0.185 & 0.251 / 0.266 / 0.317 \\
 & Keyboard & 90.63 / 90.00 / 88.13 & 5.69 / 5.83 / 6.94 & 0.152 / 0.170 / 0.203 & 0.264 / 0.289 / 0.345 \\
MCL-WiC & Apostrophe & 56.88 / 55.00 / 60.00 & 29.39 / 31.42 / 26.59 & 0.645 / 0.656 / 0.606 & 0.957 / 0.963 / 0.900 \\
 & Diacritic & 58.13 / 57.50 / 58.75 & 29.09 / 29.76 / 28.80 & 0.640 / 0.653 / 0.648 & 0.977 / 0.980 / 0.976 \\
 & Keyboard & 53.13 / 49.38 / 51.25 & 34.38 / 36.78 / 35.28 & 0.726 / 0.764 / 0.755 & 1.079 / 1.128 / 1.106 \\
\bottomrule
\end{tabular*}
\caption[]{(continued) Uncalibrated results by model, dataset, and perturbation factor. Each cell lists Original / Low / High. Acc. and ECE are percentages; Brier and NLL retain their original scales.}
\end{table*}

\begin{table*}[tp]
\centering\small\setlength{\tabcolsep}{3pt}
\renewcommand{\arraystretch}{1.08}
\addtocounter{table}{-1}
\renewcommand{\theHtable}{\arabic{table}.part3}
\begin{tabular*}{\textwidth}{@{\extracolsep{\fill}}llcccc@{}}
\toprule
Dataset & Factor & Acc. & ECE & Brier & NLL \\
\midrule
\multicolumn{6}{l}{Gemma-3-12B-it} \\
MARC & Apostrophe & 53.13 / 53.75 / 55.00 & 37.91 / 37.03 / 37.29 & 0.786 / 0.779 / 0.786 & 1.687 / 1.668 / 1.669 \\
 & Diacritic & 51.25 / 52.50 / 52.50 & 38.71 / 37.33 / 38.46 & 0.838 / 0.836 / 0.831 & 1.831 / 1.827 / 1.808 \\
 & Keyboard & 59.38 / 55.00 / 54.38 & 29.53 / 32.91 / 33.53 & 0.677 / 0.734 / 0.754 & 1.546 / 1.648 / 1.693 \\
Belebele & Apostrophe & 93.13 / 93.13 / 92.50 & 4.36 / 4.25 / 3.90 & 0.107 / 0.115 / 0.100 & 0.211 / 0.228 / 0.198 \\
 & Diacritic & 92.50 / 91.88 / 91.25 & 6.31 / 6.24 / 6.79 & 0.129 / 0.142 / 0.151 & 0.299 / 0.320 / 0.352 \\
 & Keyboard & 94.38 / 93.75 / 92.50 & 3.65 / 3.94 / 4.07 & 0.097 / 0.110 / 0.111 & 0.239 / 0.257 / 0.267 \\
FrenchNLI & Apostrophe & 78.13 / 78.13 / 76.25 & 14.23 / 15.11 / 15.40 & 0.357 / 0.369 / 0.386 & 0.688 / 0.694 / 0.731 \\
 & Diacritic & 77.50 / 76.88 / 74.38 & 14.92 / 14.70 / 17.68 & 0.380 / 0.394 / 0.414 & 0.725 / 0.769 / 0.801 \\
 & Keyboard & 76.88 / 73.75 / 70.63 & 17.02 / 18.58 / 20.03 & 0.394 / 0.416 / 0.463 & 0.777 / 0.786 / 0.823 \\
MMLU-Math & Apostrophe & 60.00 / 60.63 / 60.63 & 24.65 / 24.44 / 24.35 & 0.619 / 0.624 / 0.621 & 1.317 / 1.333 / 1.300 \\
 & Diacritic & 50.63 / 52.50 / 49.38 & 31.93 / 30.78 / 32.80 & 0.757 / 0.755 / 0.766 & 1.542 / 1.535 / 1.576 \\
 & Keyboard & 53.75 / 49.38 / 51.25 & 30.83 / 32.59 / 30.66 & 0.726 / 0.763 / 0.743 & 1.448 / 1.480 / 1.433 \\
HellaSwag & Apostrophe & 80.63 / 80.63 / 78.75 & 10.64 / 10.21 / 11.95 & 0.293 / 0.303 / 0.308 & 0.571 / 0.597 / 0.595 \\
 & Diacritic & 83.75 / 84.38 / 83.13 & 9.83 / 10.21 / 12.12 & 0.274 / 0.268 / 0.274 & 0.582 / 0.558 / 0.562 \\
 & Keyboard & 83.75 / 83.13 / 78.13 & 9.55 / 7.25 / 12.62 & 0.263 / 0.267 / 0.333 & 0.511 / 0.510 / 0.613 \\
Histoires & Apostrophe & 93.75 / 94.38 / 94.38 & 4.94 / 4.16 / 3.82 & 0.103 / 0.097 / 0.101 & 0.178 / 0.178 / 0.184 \\
 & Diacritic & 95.00 / 95.00 / 95.00 & 3.49 / 3.64 / 3.53 & 0.091 / 0.090 / 0.087 & 0.175 / 0.174 / 0.167 \\
 & Keyboard & 89.38 / 91.25 / 91.88 & 8.03 / 6.59 / 5.61 & 0.158 / 0.141 / 0.132 & 0.258 / 0.238 / 0.231 \\
MCL-WiC & Apostrophe & 76.25 / 78.13 / 78.75 & 15.87 / 13.74 / 14.05 & 0.368 / 0.347 / 0.352 & 0.597 / 0.569 / 0.576 \\
 & Diacritic & 80.00 / 80.63 / 78.75 & 15.19 / 12.58 / 13.35 & 0.343 / 0.326 / 0.334 & 0.557 / 0.526 / 0.537 \\
 & Keyboard & 70.00 / 70.63 / 69.38 & 23.45 / 22.82 / 22.67 & 0.505 / 0.499 / 0.504 & 0.820 / 0.813 / 0.815 \\
\midrule
\multicolumn{6}{l}{Gemma-3-27B-it} \\
MARC & Apostrophe & 52.50 / 51.25 / 50.00 & 41.69 / 43.01 / 43.80 & 0.866 / 0.876 / 0.886 & 2.330 / 2.319 / 2.326 \\
 & Diacritic & 49.38 / 50.00 / 50.63 & 44.01 / 43.84 / 43.90 & 0.902 / 0.894 / 0.911 & 2.437 / 2.434 / 2.483 \\
 & Keyboard & 54.38 / 51.25 / 51.88 & 38.10 / 40.62 / 40.13 & 0.813 / 0.833 / 0.838 & 2.150 / 2.174 / 2.221 \\
Belebele & Apostrophe & 93.75 / 95.00 / 94.38 & 5.05 / 4.15 / 3.64 & 0.104 / 0.087 / 0.089 & 0.234 / 0.214 / 0.225 \\
 & Diacritic & 91.88 / 91.88 / 90.63 & 7.24 / 7.14 / 8.04 & 0.156 / 0.154 / 0.173 & 0.392 / 0.396 / 0.438 \\
 & Keyboard & 93.13 / 93.13 / 94.38 & 5.58 / 5.82 / 4.92 & 0.109 / 0.117 / 0.103 & 0.247 / 0.257 / 0.234 \\
FrenchNLI & Apostrophe & 80.63 / 80.00 / 78.13 & 16.38 / 15.52 / 17.18 & 0.359 / 0.351 / 0.367 & 0.821 / 0.807 / 0.830 \\
 & Diacritic & 81.88 / 81.25 / 80.63 & 15.70 / 17.50 / 16.66 & 0.346 / 0.354 / 0.359 & 0.813 / 0.826 / 0.835 \\
 & Keyboard & 73.13 / 68.75 / 71.25 & 22.33 / 26.66 / 23.10 & 0.471 / 0.560 / 0.522 & 1.069 / 1.273 / 1.173 \\
MMLU-Math & Apostrophe & 71.25 / 70.00 / 69.38 & 19.53 / 20.58 / 21.95 & 0.477 / 0.483 / 0.498 & 1.095 / 1.104 / 1.118 \\
 & Diacritic & 69.38 / 69.38 / 70.00 & 20.79 / 19.11 / 19.46 & 0.468 / 0.454 / 0.465 & 0.896 / 0.874 / 0.890 \\
 & Keyboard & 67.50 / 70.00 / 69.38 & 22.85 / 19.05 / 20.17 & 0.518 / 0.496 / 0.525 & 1.121 / 1.089 / 1.172 \\
HellaSwag & Apostrophe & 85.00 / 84.38 / 85.63 & 11.25 / 10.87 / 9.54 & 0.252 / 0.266 / 0.247 & 0.601 / 0.600 / 0.567 \\
 & Diacritic & 88.13 / 88.75 / 88.75 & 8.26 / 8.95 / 8.34 & 0.208 / 0.207 / 0.200 & 0.514 / 0.510 / 0.495 \\
 & Keyboard & 85.00 / 80.00 / 81.88 & 11.43 / 14.66 / 11.91 & 0.259 / 0.321 / 0.290 & 0.570 / 0.683 / 0.617 \\
Histoires & Apostrophe & 94.38 / 95.00 / 95.00 & 3.80 / 3.83 / 3.24 & 0.090 / 0.079 / 0.077 & 0.179 / 0.157 / 0.164 \\
 & Diacritic & 95.00 / 95.00 / 94.38 & 4.10 / 4.22 / 4.52 & 0.084 / 0.092 / 0.091 & 0.196 / 0.208 / 0.199 \\
 & Keyboard & 95.63 / 93.75 / 93.75 & 4.35 / 4.78 / 3.02 & 0.084 / 0.096 / 0.079 & 0.175 / 0.174 / 0.137 \\
MCL-WiC & Apostrophe & 73.75 / 72.50 / 73.75 & 20.21 / 22.56 / 20.80 & 0.448 / 0.463 / 0.443 & 0.950 / 0.981 / 0.938 \\
 & Diacritic & 81.88 / 83.13 / 81.88 & 14.38 / 11.80 / 14.29 & 0.308 / 0.291 / 0.311 & 0.634 / 0.620 / 0.643 \\
 & Keyboard & 74.38 / 75.00 / 73.13 & 18.71 / 18.69 / 22.05 & 0.415 / 0.436 / 0.478 & 0.917 / 1.012 / 1.138 \\
\bottomrule
\end{tabular*}
\caption[]{(continued) Uncalibrated results by model, dataset, and perturbation factor. Each cell lists Original / Low / High. Acc. and ECE are percentages; Brier and NLL retain their original scales.}
\end{table*}

\begin{table*}[tp]
\centering
\small\setlength{\tabcolsep}{3pt}
\renewcommand{\arraystretch}{1.08}
\begin{tabular}{llcccccccc}
\toprule
Model & Intensity & Flip & Harm & Help & $\Delta$Acc. & $\Delta$Conf. & $\Delta p_{gold}$ & HCE O & HCE V \\
\midrule
Qwen3-8B & Low & 4.23 & 2.11 & 1.34 & -0.77 & -0.55 & -0.88 & 7.74 & 7.68 \\
 & High & 5.89 & 3.42 & 1.64 & -1.79 & -1.03 & -1.61 & 7.74 & 7.95 \\
Qwen3-14B & Low & 3.39 & 1.58 & 1.19 & -0.39 & -0.74 & -0.68 & 8.07 & 7.80 \\
 & High & 4.91 & 2.68 & 1.37 & -1.31 & -1.46 & -1.42 & 8.07 & 7.86 \\
Qwen3-32B & Low & 3.18 & 1.55 & 1.37 & -0.18 & -0.87 & -0.93 & 2.80 & 2.80 \\
 & High & 4.49 & 2.35 & 1.67 & -0.68 & -1.41 & -1.53 & 2.80 & 2.47 \\
Gemma-3-4B-it & Low & 7.65 & 3.51 & 2.77 & -0.74 & -0.61 & -0.79 & 24.55 & 24.29 \\
 & High & 11.52 & 5.30 & 3.96 & -1.34 & -1.10 & -1.40 & 24.55 & 23.33 \\
Gemma-3-12B-it & Low & 4.23 & 1.90 & 1.73 & -0.18 & -0.48 & -0.61 & 17.68 & 17.95 \\
 & High & 6.67 & 3.45 & 2.29 & -1.16 & -0.81 & -1.29 & 17.68 & 18.04 \\
Gemma-3-27B-it & Low & 3.75 & 1.99 & 1.40 & -0.60 & -0.22 & -0.56 & 16.64 & 16.93 \\
 & High & 5.36 & 2.68 & 2.05 & -0.63 & -0.49 & -0.84 & 16.64 & 16.88 \\
\bottomrule
\end{tabular}
\caption{Paired prediction changes. All $\Delta$ values are Variant $-$ Original. Flip and high-confidence error (HCE) rates are percentages; changes in accuracy, mean confidence (Conf.), and mean correct-answer probability ($p_{gold}$) are percentage points. O and V denote original and variant inputs.}
\label{tab:paired-changes}
\end{table*}

\subsection{Paired Prediction Changes}\label{app:paired-changes}

Prediction flips include changes between two incorrect answers. Harmful flips change a correct answer to an incorrect one; helpful flips do the reverse. Rates use the number of test triplets as their denominator. HCEs are incorrect predictions with confidence at least 0.8.

\FloatBarrier
\section{Prediction Flips and Calibration Temperatures}\label{app:confidence-details}

\textbf{Original decision margins.} Originally correct predictions are sorted by their original top-two probability margin and divided into four equally sized groups. The same groups are used for both intensities; ties are ordered by identifier.

\begin{table}[htbp]
\centering
\small\setlength{\tabcolsep}{3pt}
\renewcommand{\arraystretch}{1.08}
\begin{tabular}{lcc}
\toprule
Model & Low & High \\
\midrule
Qwen3-8B & \shortstack{91.55\\{}[84.48, 97.33]} & \shortstack{86.09\\{}[79.49, 92.17]} \\
Qwen3-14B & \shortstack{96.23\\{}[90.38, 100.00]} & \shortstack{83.33\\{}[75.55, 90.70]} \\
Qwen3-32B & \shortstack{92.31\\{}[84.09, 98.18]} & \shortstack{93.67\\{}[87.50, 98.63]} \\
Gemma-3-4B-it & \shortstack{72.88\\{}[64.66, 80.61]} & \shortstack{63.48\\{}[56.29, 70.41]} \\
Gemma-3-12B-it & \shortstack{96.88\\{}[91.94, 100.00]} & \shortstack{85.34\\{}[78.51, 91.35]} \\
Gemma-3-27B-it & \shortstack{94.03\\{}[87.88, 98.67]} & \shortstack{86.67\\{}[79.27, 93.27]} \\
\bottomrule
\end{tabular}
\caption{Share of correct-to-incorrect flips occurring in the lowest original-margin quartile (Q1). The denominator is the total number of correct-to-incorrect flips across all four quartiles. Values are percentages with pointwise 95\% percentile CIs from 10,000 bootstrap resamples over source groups, keeping the original quartile assignments fixed.}
\label{tab:q1-concentration}
\end{table}

\textbf{Calibration temperatures.} Each temperature is fitted by minimizing NLL on one input condition from the held-out triplets. Test triplets are evaluated using these fixed temperatures.

\begin{table}[htbp]
\centering
\small\setlength{\tabcolsep}{3pt}
\renewcommand{\arraystretch}{1.08}
\begin{tabular}{lccc}
\toprule
Model & Original & Low & High \\
\midrule
Qwen3-8B & 1.1496 & 1.1493 & 1.1652 \\
Qwen3-14B & 1.2223 & 1.2360 & 1.2220 \\
Qwen3-32B & 0.8184 & 0.7948 & 0.8045 \\
Gemma-3-4B-it & 2.0327 & 2.1003 & 2.0985 \\
Gemma-3-12B-it & 1.8867 & 1.8412 & 1.8106 \\
Gemma-3-27B-it & 2.3754 & 2.3778 & 2.3655 \\
\bottomrule
\end{tabular}
\caption{Temperatures fitted on the corresponding original, low-intensity, or high-intensity held-out inputs.}
\label{tab:temperatures}
\end{table}

\textbf{Deletion control.} For each variant, the deletion control removes the words modified by that variant. Comparing Variant with Deletion assesses the contribution of these words in their modified form to the correct-candidate margin. Here the correct-candidate log-score margin is the mean log score of the correct candidate minus the highest mean log score among incorrect candidates. We report Original minus Variant ($\Delta_{\text{orig-var}}$) and Variant minus Deletion ($\Delta_{\text{var-del}}$); their sum equals Original minus Deletion. Results are reported for all test triplets and the subset with correct-to-incorrect flips.

\begin{table*}[!t]
\centering
\small\setlength{\tabcolsep}{3pt}
\renewcommand{\arraystretch}{1.08}
\begin{tabular}{llccccccc}
\toprule
 & & \multicolumn{3}{c}{All test triplets} & \multicolumn{4}{c}{Correct to incorrect} \\
\cmidrule(lr){3-5}\cmidrule(lr){6-9}
Model & Intensity & $\Delta_{\text{orig-var}}$ & $\Delta_{\text{var-del}}$ & 95\% CI & $n$ & $\Delta_{\text{orig-var}}$ & $\Delta_{\text{var-del}}$ & 95\% CI \\
\midrule
Qwen3-8B & Low & 0.069 & 0.202 & [0.174, 0.230] & 71 & 0.848 & -0.083 & [-0.305, 0.137] \\
 & High & 0.111 & 0.396 & [0.357, 0.435] & 115 & 1.137 & -0.116 & [-0.323, 0.092] \\
Qwen3-14B & Low & 0.090 & 0.304 & [0.269, 0.338] & 53 & 1.107 & -0.248 & [-0.631, 0.142] \\
 & High & 0.153 & 0.588 & [0.543, 0.636] & 90 & 1.461 & 0.242 & [-0.026, 0.505] \\
Qwen3-32B & Low & 0.063 & 0.184 & [0.160, 0.207] & 52 & 0.766 & -0.142 & [-0.351, 0.059] \\
 & High & 0.100 & 0.388 & [0.357, 0.419] & 79 & 0.870 & 0.126 & [-0.041, 0.293] \\
Gemma-3-4B-it & Low & 0.066 & 0.154 & [0.107, 0.202] & 118 & 3.135 & -1.066 & [-1.453, -0.703] \\
 & High & 0.111 & 0.309 & [0.248, 0.369] & 178 & 3.544 & -1.522 & [-1.830, -1.218] \\
Gemma-3-12B-it & Low & 0.135 & 0.286 & [0.234, 0.338] & 64 & 2.764 & -0.960 & [-1.512, -0.402] \\
 & High & 0.222 & 0.608 & [0.535, 0.680] & 116 & 3.352 & -1.021 & [-1.477, -0.585] \\
Gemma-3-27B-it & Low & 0.155 & 0.408 & [0.345, 0.472] & 67 & 2.775 & -0.486 & [-1.139, 0.158] \\
 & High & 0.220 & 0.854 & [0.771, 0.936] & 90 & 3.812 & -0.566 & [-1.149, 0.011] \\
\bottomrule
\end{tabular}
\caption{Deletion-control margin differences. Subscripts indicate the order of subtraction: orig denotes Original, var denotes Variant, and del denotes Deletion. Each CI refers to $\Delta_{\text{var-del}}$ and is a pointwise 95\% percentile interval from 10,000 paired bootstrap resamples over source groups.}
\label{tab:deletion}
\end{table*}

\begin{table*}[!t]
\centering
\small\setlength{\tabcolsep}{3pt}
\renewcommand{\arraystretch}{1.08}
\begin{tabular}{lccccccccc}
\toprule
 & \multicolumn{3}{c}{Original} & \multicolumn{3}{c}{Low} & \multicolumn{3}{c}{High} \\
\cmidrule(lr){2-4}\cmidrule(lr){5-7}\cmidrule(lr){8-10}
Model & $\Delta$Acc. & Repairs & Harms & $\Delta$Acc. & Repairs & Harms & $\Delta$Acc. & Repairs & Harms \\
\midrule
Qwen3-8B & +0.03 & 5 & 4 & +0.15 & 54 & 49 & +0.92 & 83 & 52 \\
Qwen3-14B & -0.03 & 2 & 3 & +0.21 & 45 & 38 & +0.77 & 73 & 47 \\
Qwen3-32B & -0.06 & 2 & 4 & +0.18 & 42 & 36 & +0.33 & 64 & 53 \\
Gemma-3-4B-it & -0.03 & 9 & 10 & +0.03 & 93 & 92 & +0.63 & 146 & 125 \\
Gemma-3-12B-it & -0.03 & 3 & 4 & -0.15 & 40 & 45 & +0.21 & 87 & 80 \\
Gemma-3-27B-it & +0.00 & 3 & 3 & +0.48 & 50 & 34 & +0.24 & 64 & 56 \\
\bottomrule
\end{tabular}
\caption{Changes after input restoration on 3,360 test triplets, reported separately for each model and input condition. $\Delta$Acc. is restored minus raw accuracy, in percentage points; Repairs and Harms are counts.}
\label{tab:restoration-full}
\end{table*}

\section{Calibration and Restoration Results}\label{app:interventions}

This appendix supplements Section~\ref{sec:calibration-restoration} with task-level calibration findings and original-input controls.

\subsection{Calibration Transfer}\label{app:calibration-transfer}

Each model uses its temperature fitted on the original inputs of the held-out triplets. ECE is recomputed from pooled predictions in 10 equal-width bins. Calibration and Combined use the same original temperature, applied after restoration for Combined; no temperature is fitted on restored inputs.

Although global calibration improves pooled NLL, its benefits vary across datasets. Temperatures fitted separately on each dataset's original held-out inputs reverse most dataset-level NLL increases at both perturbation intensities.

\subsection{Input Restoration and Original-Input Control}\label{app:restoration}

Repairs count incorrect predictions that become correct after restoration; harms count correct predictions that become incorrect. Both counts use the input before restoration as their reference. The original-input control applies the same restoration procedure to original test inputs.

\section{Paired Effect Estimates}
\label{app:effect-intervals}
Tables~\ref{tab:overall-intervals}--\ref{tab:combined-intervals} report differences and pointwise 95\% paired bootstrap CIs for the main results and intervention comparisons. Each comparison uses the same test triplets, matching predictions across conditions or methods; sampling details are given in Appendix~\ref{app:statistics}.
\begin{table*}[tp]
\centering\small\setlength{\tabcolsep}{4pt}
\begin{tabular*}{\textwidth}{@{\extracolsep{\fill}}llcccc@{}}
\toprule
 & & \multicolumn{2}{c}{Low} & \multicolumn{2}{c}{High} \\
\cmidrule(lr){3-4}\cmidrule(lr){5-6}
Model & Metric & Difference [95\% CI] & Adjusted $p$ & Difference [95\% CI] & Adjusted $p$ \\
\midrule
Qwen3-8B & Acc. & \shortstack{$-0.77$\\$[-1.40,-0.15]$} & $0.5269$ & \shortstack{$-1.79$\\$[-2.53,-1.05]$} & $0.0048$ \\
 & ECE & \shortstack{$0.21$\\$[-0.42,0.96]$} & $1.0000$ & \shortstack{$0.94$\\$[0.03,1.58]$} & $0.6859$ \\
 & Brier & \shortstack{$0.0094$\\$[0.0058,0.0133]$} & $0.0048$ & \shortstack{$0.0183$\\$[0.0131,0.0235]$} & $0.0048$ \\
 & NLL & \shortstack{$0.0141$\\$[0.0085,0.0199]$} & $0.0048$ & \shortstack{$0.0271$\\$[0.0190,0.0352]$} & $0.0048$ \\
\midrule
Qwen3-14B & Acc. & \shortstack{$-0.39$\\$[-0.95,0.21]$} & $1.0000$ & \shortstack{$-1.31$\\$[-1.99,-0.66]$} & $0.0048$ \\
 & ECE & \shortstack{$-0.67$\\$[-1.25,-0.003]$} & $0.9593$ & \shortstack{$-0.46$\\$[-1.24,0.25]$} & $1.0000$ \\
 & Brier & \shortstack{$0.0037$\\$[0.0001,0.0075]$} & $1.0000$ & \shortstack{$0.0095$\\$[0.0044,0.0147]$} & $0.0048$ \\
 & NLL & \shortstack{$0.0079$\\$[0.0022,0.0138]$} & $0.2268$ & \shortstack{$0.0180$\\$[0.0096,0.0265]$} & $0.0048$ \\
\midrule
Qwen3-32B & Acc. & \shortstack{$-0.18$\\$[-0.77,0.41]$} & $1.0000$ & \shortstack{$-0.68$\\$[-1.37,0.00]$} & $1.0000$ \\
 & ECE & \shortstack{$0.43$\\$[-0.42,1.12]$} & $1.0000$ & \shortstack{$0.80$\\$[-0.28,1.46]$} & $1.0000$ \\
 & Brier & \shortstack{$0.0071$\\$[0.0040,0.0103]$} & $0.0074$ & \shortstack{$0.0128$\\$[0.0089,0.0169]$} & $0.0048$ \\
 & NLL & \shortstack{$0.0143$\\$[0.0097,0.0190]$} & $0.0048$ & \shortstack{$0.0235$\\$[0.0174,0.0296]$} & $0.0048$ \\
\midrule
Gemma-3-4B-it & Acc. & \shortstack{$-0.74$\\$[-1.58,0.09]$} & $1.0000$ & \shortstack{$-1.34$\\$[-2.35,-0.33]$} & $0.3102$ \\
 & ECE & \shortstack{$0.13$\\$[-0.73,1.01]$} & $1.0000$ & \shortstack{$0.24$\\$[-0.80,1.27]$} & $1.0000$ \\
 & Brier & \shortstack{$0.0063$\\$[-0.0038,0.0165]$} & $1.0000$ & \shortstack{$0.0108$\\$[-0.0024,0.0236]$} & $1.0000$ \\
 & NLL & \shortstack{$0.0039$\\$[-0.0120,0.0201]$} & $1.0000$ & \shortstack{$0.0065$\\$[-0.0146,0.0272]$} & $1.0000$ \\
\midrule
Gemma-3-12B-it & Acc. & \shortstack{$-0.18$\\$[-0.83,0.48]$} & $1.0000$ & \shortstack{$-1.16$\\$[-1.97,-0.33]$} & $0.2268$ \\
 & ECE & \shortstack{$-0.24$\\$[-0.87,0.49]$} & $1.0000$ & \shortstack{$0.35$\\$[-0.47,1.18]$} & $1.0000$ \\
 & Brier & \shortstack{$0.0053$\\$[-0.0023,0.0130]$} & $1.0000$ & \shortstack{$0.0137$\\$[0.0039,0.0236]$} & $0.2652$ \\
 & NLL & \shortstack{$0.0071$\\$[-0.0055,0.0198]$} & $1.0000$ & \shortstack{$0.0177$\\$[0.0006,0.0350]$} & $1.0000$ \\
\midrule
Gemma-3-27B-it & Acc. & \shortstack{$-0.60$\\$[-1.20,0.03]$} & $1.0000$ & \shortstack{$-0.63$\\$[-1.36,0.09]$} & $1.0000$ \\
 & ECE & \shortstack{$0.25$\\$[-0.38,0.88]$} & $1.0000$ & \shortstack{$0.19$\\$[-0.56,0.92]$} & $1.0000$ \\
 & Brier & \shortstack{$0.0083$\\$[0.0013,0.0154]$} & $0.6553$ & \shortstack{$0.0103$\\$[0.0012,0.0196]$} & $0.8261$ \\
 & NLL & \shortstack{$0.0176$\\$[0.0037,0.0323]$} & $0.5269$ & \shortstack{$0.0238$\\$[0.0052,0.0427]$} & $0.3872$ \\
\bottomrule
\end{tabular*}
\caption{Metric differences (Variant $-$ Original) corresponding to Table~\ref{tab:main-results}. Acc. and ECE differences are in percentage points; Brier and NLL retain their original scales. CIs are pointwise 95\% percentile intervals from paired bootstrap resampling; $p$ values are Holm-adjusted.}
\label{tab:overall-intervals}
\end{table*}

\begin{table*}[tp]
\centering\small
\begin{tabular*}{\textwidth}{@{\extracolsep{\fill}}llcccc@{}}
\toprule
\multicolumn{6}{l}{(a) Calibration and restoration relative to raw variants} \\
\midrule
 & & \multicolumn{3}{c}{Calibration $-$ Raw} & Restoration $-$ Raw \\
\cmidrule(lr){3-5}\cmidrule(lr){6-6}
Model & Intensity & $\Delta$ECE & $\Delta$Brier & $\Delta$NLL & $\Delta$Acc. \\
\midrule
Qwen3-8B & Low & \shortstack{$-2.48$\\$[-3.24,-1.07]$} & \shortstack{$-0.0034$\\$[-0.0050,-0.0018]$} & \shortstack{$-0.0057$\\$[-0.0096,-0.0019]$} & \shortstack{$0.15$\\$[-0.42,0.75]$} \\
 & High & \shortstack{$-3.18$\\$[-3.55,-1.71]$} & \shortstack{$-0.0038$\\$[-0.0054,-0.0022]$} & \shortstack{$-0.0062$\\$[-0.0101,-0.0024]$} & \shortstack{$0.92$\\$[0.24,1.58]$} \\
\addlinespace
Qwen3-14B & Low & \shortstack{$-3.75$\\$[-4.25,-2.38]$} & \shortstack{$-0.0067$\\$[-0.0087,-0.0047]$} & \shortstack{$-0.0144$\\$[-0.0200,-0.0091]$} & \shortstack{$0.21$\\$[-0.33,0.74]$} \\
 & High & \shortstack{$-3.46$\\$[-4.21,-2.40]$} & \shortstack{$-0.0064$\\$[-0.0084,-0.0044]$} & \shortstack{$-0.0136$\\$[-0.0191,-0.0083]$} & \shortstack{$0.77$\\$[0.15,1.42]$} \\
\addlinespace
Qwen3-32B & Low & \shortstack{$-2.27$\\$[-3.80,-0.72]$} & \shortstack{$-0.0021$\\$[-0.0040,-0.0002]$} & \shortstack{$-0.0120$\\$[-0.0163,-0.0077]$} & \shortstack{$0.18$\\$[-0.33,0.69]$} \\
 & High & \shortstack{$-3.07$\\$[-4.28,-1.05]$} & \shortstack{$-0.0020$\\$[-0.0039,-0.0001]$} & \shortstack{$-0.0122$\\$[-0.0165,-0.0079]$} & \shortstack{$0.33$\\$[-0.30,0.98]$} \\
\addlinespace
Gemma-3-4B-it & Low & \shortstack{$-18.43$\\$[-19.48,-16.74]$} & \shortstack{$-0.0915$\\$[-0.1019,-0.0815]$} & \shortstack{$-0.2063$\\$[-0.2296,-0.1836]$} & \shortstack{$0.03$\\$[-0.75,0.82]$} \\
 & High & \shortstack{$-18.64$\\$[-19.73,-17.00]$} & \shortstack{$-0.0917$\\$[-0.1018,-0.0815]$} & \shortstack{$-0.2020$\\$[-0.2251,-0.1796]$} & \shortstack{$0.63$\\$[-0.32,1.57]$} \\
\addlinespace
Gemma-3-12B-it & Low & \shortstack{$-10.52$\\$[-11.47,-9.15]$} & \shortstack{$-0.0541$\\$[-0.0612,-0.0471]$} & \shortstack{$-0.1619$\\$[-0.1846,-0.1398]$} & \shortstack{$-0.15$\\$[-0.69,0.39]$} \\
 & High & \shortstack{$-10.69$\\$[-11.52,-9.32]$} & \shortstack{$-0.0561$\\$[-0.0631,-0.0490]$} & \shortstack{$-0.1615$\\$[-0.1837,-0.1399]$} & \shortstack{$0.21$\\$[-0.54,0.96]$} \\
\addlinespace
Gemma-3-27B-it & Low & \shortstack{$-12.95$\\$[-14.44,-11.33]$} & \shortstack{$-0.0573$\\$[-0.0656,-0.0493]$} & \shortstack{$-0.2929$\\$[-0.3302,-0.2564]$} & \shortstack{$0.48$\\$[-0.06,0.99]$} \\
 & High & \shortstack{$-13.52$\\$[-14.80,-11.79]$} & \shortstack{$-0.0556$\\$[-0.0637,-0.0477]$} & \shortstack{$-0.2918$\\$[-0.3294,-0.2558]$} & \shortstack{$0.24$\\$[-0.39,0.86]$} \\
\midrule
\multicolumn{6}{l}{(b) Combined relative to each intervention alone} \\
\midrule
 & & \multicolumn{2}{c}{Combined $-$ Calibration} & \multicolumn{2}{c}{Combined $-$ Restoration} \\
\cmidrule(lr){3-4}\cmidrule(lr){5-6}
Model & Intensity & $\Delta$Brier & $\Delta$NLL & $\Delta$Brier & $\Delta$NLL \\
\midrule
Qwen3-8B & Low & \shortstack{$-0.0048$\\$[-0.0077,-0.0021]$} & \shortstack{$-0.0076$\\$[-0.0119,-0.0034]$} & \shortstack{$-0.0034$\\$[-0.0050,-0.0018]$} & \shortstack{$-0.0058$\\$[-0.0097,-0.0020]$} \\
 & High & \shortstack{$-0.0097$\\$[-0.0136,-0.0058]$} & \shortstack{$-0.0150$\\$[-0.0207,-0.0091]$} & \shortstack{$-0.0034$\\$[-0.0050,-0.0018]$} & \shortstack{$-0.0056$\\$[-0.0095,-0.0018]$} \\
\addlinespace
Qwen3-14B & Low & \shortstack{$-0.0021$\\$[-0.0049,0.0006]$} & \shortstack{$-0.0054$\\$[-0.0095,-0.0015]$} & \shortstack{$-0.0073$\\$[-0.0093,-0.0053]$} & \shortstack{$-0.0156$\\$[-0.0213,-0.0102]$} \\
 & High & \shortstack{$-0.0055$\\$[-0.0093,-0.0016]$} & \shortstack{$-0.0100$\\$[-0.0155,-0.0044]$} & \shortstack{$-0.0070$\\$[-0.0090,-0.0050]$} & \shortstack{$-0.0152$\\$[-0.0210,-0.0098]$} \\
\addlinespace
Qwen3-32B & Low & \shortstack{$-0.0050$\\$[-0.0083,-0.0020]$} & \shortstack{$-0.0090$\\$[-0.0139,-0.0043]$} & \shortstack{$-0.0018$\\$[-0.0037,0.0001]$} & \shortstack{$-0.0116$\\$[-0.0159,-0.0072]$} \\
 & High & \shortstack{$-0.0078$\\$[-0.0119,-0.0037]$} & \shortstack{$-0.0138$\\$[-0.0201,-0.0076]$} & \shortstack{$-0.0018$\\$[-0.0037,0.0001]$} & \shortstack{$-0.0120$\\$[-0.0162,-0.0076]$} \\
\addlinespace
Gemma-3-4B-it & Low & \shortstack{$-0.0012$\\$[-0.0063,0.0040]$} & \shortstack{$-0.0029$\\$[-0.0099,0.0042]$} & \shortstack{$-0.0949$\\$[-0.1050,-0.0849]$} & \shortstack{$-0.2142$\\$[-0.2378,-0.1914]$} \\
 & High & \shortstack{$-0.0041$\\$[-0.0106,0.0024]$} & \shortstack{$-0.0071$\\$[-0.0160,0.0020]$} & \shortstack{$-0.0949$\\$[-0.1052,-0.0848]$} & \shortstack{$-0.2113$\\$[-0.2349,-0.1886]$} \\
\addlinespace
Gemma-3-12B-it & Low & \shortstack{$-0.0021$\\$[-0.0057,0.0017]$} & \shortstack{$-0.0057$\\$[-0.0109,-0.0004]$} & \shortstack{$-0.0557$\\$[-0.0628,-0.0488]$} & \shortstack{$-0.1672$\\$[-0.1903,-0.1452]$} \\
 & High & \shortstack{$-0.0049$\\$[-0.0102,0.0004]$} & \shortstack{$-0.0106$\\$[-0.0179,-0.0032]$} & \shortstack{$-0.0570$\\$[-0.0642,-0.0500]$} & \shortstack{$-0.1670$\\$[-0.1897,-0.1452]$} \\
\addlinespace
Gemma-3-27B-it & Low & \shortstack{$-0.0062$\\$[-0.0099,-0.0027]$} & \shortstack{$-0.0107$\\$[-0.0162,-0.0056]$} & \shortstack{$-0.0556$\\$[-0.0637,-0.0477]$} & \shortstack{$-0.2861$\\$[-0.3231,-0.2496]$} \\
 & High & \shortstack{$-0.0071$\\$[-0.0118,-0.0025]$} & \shortstack{$-0.0134$\\$[-0.0201,-0.0068]$} & \shortstack{$-0.0554$\\$[-0.0636,-0.0477]$} & \shortstack{$-0.2869$\\$[-0.3238,-0.2507]$} \\
\bottomrule
\end{tabular*}
\caption{Intervention differences and pointwise 95\% paired bootstrap CIs. Each cell gives the difference followed by its CI; column headings specify the order of subtraction. Acc. and ECE differences are in percentage points; Brier and NLL retain their original scales. Negative probability-score differences and positive accuracy differences indicate improvement. All calibrated scores use the temperature fitted on original held-out inputs.}
\label{tab:intervention-intervals}
\label{tab:combined-intervals}
\end{table*}

\end{document}